\documentclass[letterpaper, 10 pt, journal, twoside]{IEEEtran} 
\IEEEoverridecommandlockouts                              

\usepackage[dvipsnames]{xcolor}
\definecolor{dark-green}{RGB}{12,80,12}

\usepackage{url}
\usepackage{dsfont}
\usepackage{amsmath}
\usepackage{amssymb}
\usepackage[mathscr]{euscript}
\DeclareMathOperator{\E}{\mathbb{E}}
\newcommand{\expnumber}[2]{{#1}\mathrm{e}{#2}}
\usepackage{booktabs}
\usepackage{xspace}
\usepackage{caption}
\usepackage{graphicx} 
\usepackage{color}
\usepackage{siunitx}
\usepackage[hang,flushmargin]{footmisc}
\usepackage{microtype}
\newcommand{\secref}[1]{Sec.~\ref{#1}}
\renewcommand{\eqref}[1]{Eq.~(\ref{#1})}
\newcommand{\figref}[1]{Fig.~\ref{#1}}
\newcommand{\tabref}[1]{Tab.~\ref{#1}}

\newcommand{\blue}[1]{{#1}}
\newcommand{\green}[1]{{#1}}
\usepackage{tabularx}
\usepackage{colortbl} 
\newcolumntype{Y}{>{\centering\arraybackslash}X}
\newcolumntype{Z}{>{\raggedleft\arraybackslash}X}
\usepackage[normalem]{ulem}

\renewcommand{\sout}[1]{}

\DeclareSIUnit{\rad}{rad}

\title{Learning Kinematic Feasibility for Mobile Manipulation through Deep Reinforcement Learning}
\author{Daniel Honerkamp, Tim Welschehold, and Abhinav Valada
\thanks{\hspace{5pt}Manuscript received: January, 14, 2021; Revised April, 21, 2021; Accepted June, 14, 2021.}%
\thanks{\hspace{5pt}This paper was recommended for publication by Editor Markus Vincze upon evaluation of the Associate Editor and Reviewers' comments.
This work was supported by the European Union’s Horizon 2020 research and innovation program under grant
agreement No 871449-OpenDR and a research grant from Eva Mayr-Stihl Stiftung.}%
\thanks{\hspace{5pt}All authors are with the Department of Computer Science, University of Freiburg, Germany {\tt\footnotesize honerkamp@cs.uni-freiburg.de}}%
\thanks{\hspace{5pt}Digital Object Identifier (DOI): see top of this page.}
\thanks{\hspace{5pt}© 2021 IEEE.  Personal use of this material is permitted.  Permission from IEEE must be obtained for all other uses, in any current or future media, including reprinting/republishing this material for advertising or promotional purposes, creating new collective works, for resale or redistribution to servers or lists, or reuse of any copyrighted component of this work in other works.}
}

\begin{document}

\maketitle
\begin{abstract}
Mobile manipulation tasks remain one of the critical challenges for the widespread adoption of autonomous robots in both service and industrial scenarios.
While planning approaches are good at generating feasible whole-body robot trajectories, they struggle with dynamic environments as well as the incorporation of constraints given by the task and the environment. On the other hand, dynamic motion models in the action space struggle with generating kinematically feasible trajectories for mobile manipulation actions. We propose a deep reinforcement learning approach to learn feasible dynamic motions for a mobile base while the end-effector follows a trajectory in task space generated by an arbitrary system to fulfill the task at hand. This modular formulation has several benefits: it enables us to readily transform a broad range of end-effector motions into mobile applications, it allows us to use the kinematic feasibility of the end-effector trajectory as a dense reward signal and its modular formulation allows it to generalise to unseen end-effector motions at test time. We demonstrate the capabilities of our approach on multiple mobile robot platforms with different kinematic abilities and different types of wheeled platforms in extensive simulated as well as real-world experiments. 
\end{abstract}


\section{Introduction}
\sout{In recent years, several approaches that have been proposed to improve the capabilities of robotic platforms in both industrial and dynamic domestic environments have achieved impressive results~\cite{bagnell12integrated, blomqvist2020go, stuckler2016mobile, valada2016convoluted}.} 
\IEEEPARstart{M}{obile} manipulation is a key research area on the journey to both autonomous household assistants as well as flexible automation processes and warehouse logistics. Although impressive results have been achieved over the last years \cite{bagnell12integrated, blomqvist2020go, stuckler2016mobile, xia2020relmogen}, there remain multiple unsolved research problems. One of the major ones being that\sout{However,} most \blue{current}\sout{of these} approaches separate navigation and manipulation due to the difficulties in planning the joint movement of the robot base and its end-effector (EE). \blue{This restricts the range of tasks that can be solved and constrains the overall efficiency that can be achieved.} Typically, the tasks that a robot is expected to perform are linked to conditions in the task space, such as poses at which handled objects can be grasped, orientation that objects should maintain or entire trajectories that must be followed. While there are techniques to position a manipulator to fulfill various task constraints with respect to the kinematics of the robot, based on inverse reachability maps (IRM)~\cite{8206531}, performing such tasks while moving the base still remains an unsolved problem. 

Classical planning approaches circumvent kinematic issues implicitly by exploring paths in the configuration space of the robot~\cite{6630792}. However, this creates a number of new problems. First, the constraints must be transferred from the task space to the robot specific configuration space, requiring expert knowledge on the task, the robot and the environment. Furthermore, the execution of pre-planned configuration space movements in dynamic environments is challenging as minor errors in the execution of poses in the configuration space can lead to large deviations in the task space. Moreover, adjustments to the movement might be necessary due to changes in the dynamic scene which requires complete re-planning in the configuration space.

\begin{figure}
\centering
\includegraphics[width=0.35\textwidth,trim={0.1cm 1.0cm 0.1cm 2.2cm},clip,angle =0]{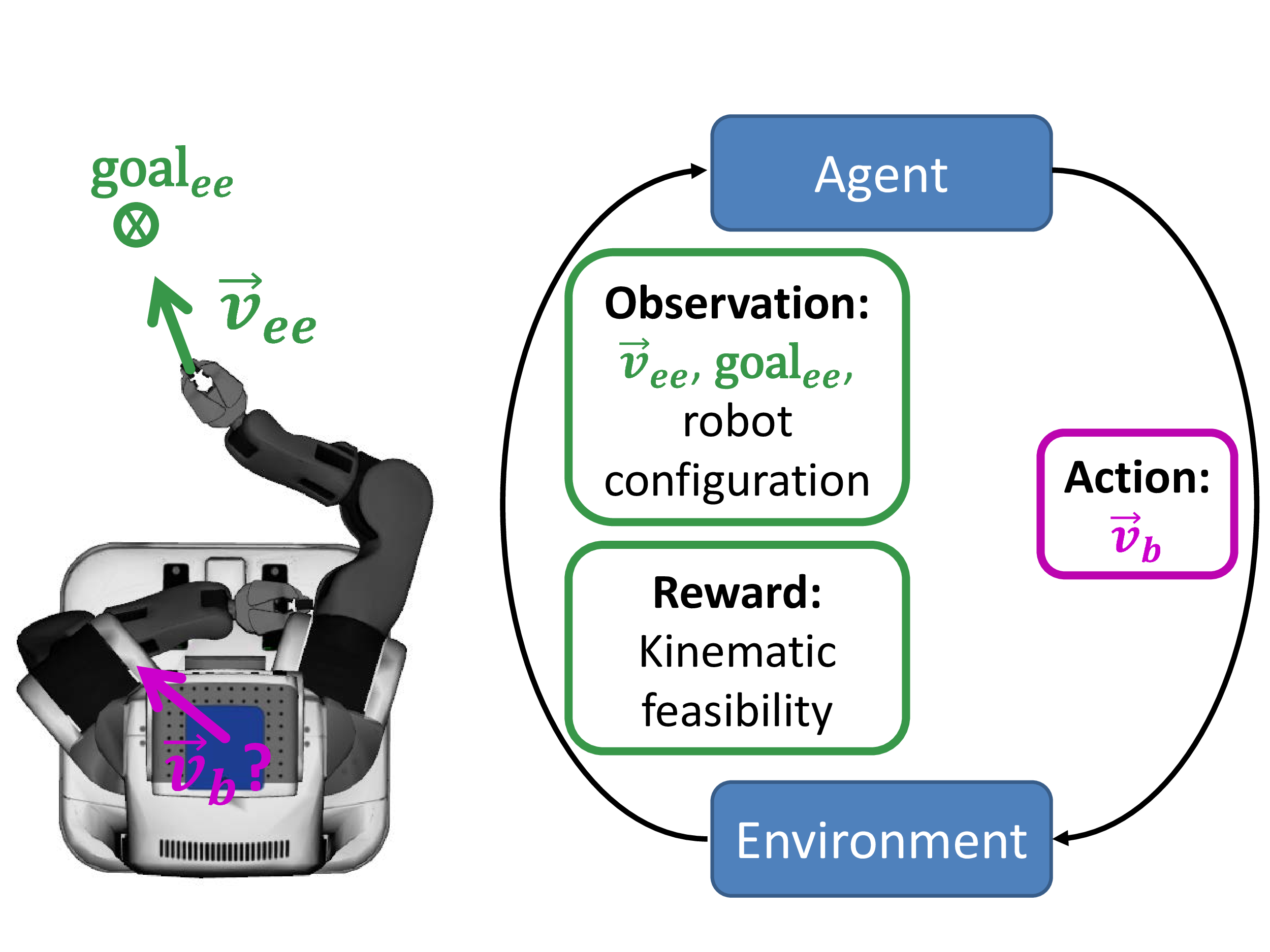}
\caption{Given the robot configuration, a velocity for the end-effector $\Vec{v}_{ee}$ and the desired goal for the end-effector, the robot learns a corresponding base velocity $\Vec{v}_{b}$ in a reinforcement learning setting to maintain kinematic feasibility throughout the motion execution.}
\label{fig:intro}
\end{figure}

In this paper, we present a method to generate kinematically feasible movement for the base of a mobile robot while its end-effector executes motions generated by an arbitrary system in the task space to perform a certain action. 
This decomposes the task into generation of trajectories for the end-effector, which is typically defined by the task constraints, and the robot base, which should handle kinematic constraints and collision avoidance. This separation is beneficial for many robotic applications. First, it results in high modularity where action models for the end-effector can easily be shared among robots with different kinematic properties. Therefore, there is no need to adapt the behavior of the end-effector for different robots as the kinematic constraints are handled entirely by the base motion control. Second, if the learned base policy is able to generalise to arbitrary end-effector motions, the same base policy can be used on new unseen tasks without expensive retraining. Lastly, mobile manipulation tasks are complex, long horizon problems that can be difficult to learn with sparse rewards only. 
We show that we can directly leverage the kinematic feasibility as a dense reward signal across platforms, alleviating the need for extensive reward shaping.

In prior work~\cite{twelsche18iros}, we addressed the kinematic feasibility of a joint base and end-effector motion by treating the inverse reachability constraint as an obstacle avoidance problem. While the approach achieves good results in mobile manipulation actions on a PR2 robot, it requires substantial robot specific design and the approximations made to model the inverse reachable space restricts the movement of the robot further than necessary. Instead of explicitly modeling the kinematic abilities, we propose to directly learn a feasible motion of the robot base respecting a given motion of the end-effector and show that the learned policy strongly outperforms these approximations. We illustrate our envisioned system in \figref{fig:intro}.

In summary, we make the following main contributions:
\begin{enumerate}
    \item We formulate \green{the fulfillment of} kinematic feasibility constraints \green{in} mobile manipulation tasks as a reinforcement learning problem. 
    \item We design multiple environments for different robots with considerably different kinematic abilities and varying driving modes.
    \item We present extensive simulated and real world experiments which demonstrate that our approach \blue{itself} generalises across diverse robotic platforms \blue{while the platform specific trained models generalise to} seen and unseen end-effector motions.
    \item We make the source code, models and videos publicly available at \url{http://rl.uni-freiburg.de/research/kinematic-feasibility-rl}.
\end{enumerate}

\section{Related Work}

In general, there are two distinct methods to ensure kinematic feasibility in mobile manipulation tasks. On one hand, planning frameworks can be used to plan trajectories for the robot in joint space and thereby only explore kinematically feasible paths~\cite{6630792,7759547}. On the other hand, inverse reachability maps~\cite{Vahrenkamp2013} can be used to seek good positioning for the robot base given the task constraints~\cite{8206531}. While combinations of the two methods exist~\cite{6907099}, it remains a hard problem to integrate kinematic feasibility constraints in task space mobile motion planning.  

In this context, Welschehold~\textit{et~al.}~\cite{twelsche18iros} propose a geometric description of the inverse reachability and address the kinematic feasibility of an arbitrary gripper trajectory as an obstacle avoidance problem. They first approximate feasible base poses relative to the end-effector resulting in a bounded region for the base. They then analytically modulate the base velocity to stay within feasible regions and orientations. Although their approach performs well on a real-world PR2 robot, the approximations require expert knowledge and do not easily generalise to different platforms. 
On a high level, we use a conceptually similar setup of given EE-motions and base control. We do not impose any geometric constraints on the allowed base poses by using reinforcement learning (RL) to directly learn the base velocities. We also show that our approach generalises to different robots without the need for robot-specific expert knowledge\blue{, i.e. can directly be used to train agents for each platform}. We directly compare with \cite{twelsche18iros} in our experiments.


While RL has shown substantial promise in manipulation tasks, it has only recently been incorporated into mobile manipulation tasks. RelMoGen proposes to learn high-level subgoals through RL to simplify the exploration problem~\cite{xia2020relmogen}. It focuses on tasks in which the exact gripper trajectories are not relevant and learns to either move the gripper or the base. In contrast, we are explicitly interested in task specific end-effector trajectories and use RL to to ensure kinematic feasibility of conjoint end-effector and base motions. 

Kindle~\textit{et~al.}~\cite{kindle2020wholebody} use RL to learn both base and end-effector movements end-to-end but they restrict the arm to lie on a plane parallel to the ground. Using a handcrafted reward with numerous hyperparameters, they demonstrate navigation in a hallway. Similarly, Wang~\textit{et~al.}~\cite{wang20} solve a mobile picking task by learning to jointly control both the base of the robot and its end-effector. While these works focus on learning a policy for one specific task, we address a more general problem of maintaining kinematic feasibility in arbitrary mobile manipulation actions. By introducing an arbitrary end-effector motion planner, we decouple the RL agent's behaviour from the exact task that the end-effector performs. This enables us to use kinematic feasibility as the sole reward signal and thereby generalises to different tasks with different motion constraints.
Recently the use of RL has also been explored to calculate forward and inverse kinematics of complex many-joint robot arms~\cite{10.1007/978-3-319-44778-0_18,10.1145/3351180.3351199}.
While such approaches are interesting for robot system with a large number of degrees of freedom (up to $40$-dof~\cite{10.1145/3351180.3351199}), in our applications ($7$-dof in robot arms) the inverse kinematics can still be solved numerically. 


Goal conditional RL~\cite{Kaelbling93learningto, schaul15} takes both the current state and a goal as input to predict the actions to arrive at this goal. Hierarchical methods~\cite{sutton1999between, bacon2017option, kaelbling1993hierarchical} commonly reduce the complexity of long-horizon tasks by splitting it into a subgoal proposal and goal-conditional policy.
Li~\textit{et~al.}~\cite{Li2019HRL4INHR} adapt hierarchical RL to mobile manipulation by incorporating an additional high-level action that restricts the low-level action space to either the base, arm or both, and demonstrate its ability to reach goals behind closed doors. While it is more-data efficient than previous hierarchical approaches, it still has to deal with the added complexity of the non-stationarity between different levels, and the additional parameters and rewards to learn both levels. In contrast, we reduce the complexity of the task by assuming a given planner for a subset of the agent (the gripper) and learn to control the remaining degrees of freedom (the base) to achieve the generated end-effector motions. This shifts the burden from the end-effector planner to the base policy and allows us to use simple and general methods to generate the end-effector motions.

\section{Learning Kinematically Feasible Robot~Base~Motions}\label{sec:approach}
\begin{figure}
\centering
\includegraphics[width=0.485\textwidth,trim={0.0cm 0cm 0cm 0cm},clip,angle =0]{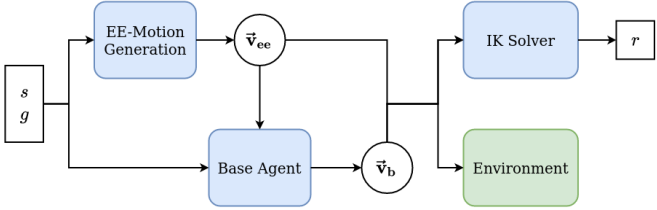}
\caption{We decompose mobile manipulation tasks into two components: an end-effector (EE) \sout{planner}\blue{motion generation} and a conditional RL agent that controls the base velocities. This enables us to produce a dense reward solely based on the kinematic feasibility, computed by a standard inverse kinematics solver.}
\label{fig:rl_scheme}
\end{figure}

Mobile manipulation tasks require complex trajectories in the conjoint space of arm and base over long horizons. We decompose the problem into a given, arbitrary motion for the end-effector and a learned base policy. This allows us to readily transform end-effector motions into mobile applications and to strongly reduce the burden on the end-effector \sout{planner}\blue{motion generation} which can now be reduced to a fairly simple system. We then formulate this as a goal-conditional RL problem and show that we can leverage kinematic feasibility as a simple, dense reward signal instead of relying on either sparse or extensively shaped rewards. Our proposed system consists of three main components: a \sout{planner}\blue{motion generator} for the end-effector, a learned RL policy for the base and a standard inverse kinematics (IK) solver for the manipulator arm that provides us with the rewards. An overview of the system is shown in \figref{fig:rl_scheme}.

\subsection{End-Effector Trajectory Generation}

To generate end-effector \blue{motions}, we assume access to an arbitrary \sout{planner}\blue{motion generator} that takes as input the current end-effector pose and a goal state $g$ \blue{in a fixed map frame}, specified by an end-effector pose in cartesian space. At every time step, this \sout{planner}\blue{generator} then outputs the next velocity command $\mathbf{v}_{ee}$ for the end-effector. During training we pass the last \sout{\textit{planned}}\blue{\textit{desired}} instead of the current EE pose to the \sout{planner}\blue{generator} as to prevent the RL agent from influencing the shape of the overall EE-trajectory. At test time the EE-\sout{planner}\blue{motion generator} can be replaced by an arbitrary system \sout{, including both closed and open loop approaches}.

\subsection{Learning Robot Base Trajectories}\label{sec:RL}

Given an end-effector motion dynamic, our goal is to ensure that the resulting EE-poses remain kinematically feasible at every time step. We can formulate this as a goal-conditioned reinforcement learning problem. We define a finite-horizon Markov decision process (MDP) $\mathcal{M} = (\mathcal{S}, \mathcal{A}, P, r, \gamma)$ with state and action spaces $\mathcal{S}$ and $\mathcal{A}$, transition dynamics $P(s_{t+1} |s_t, a_t)$, a reward function $r$ and a discount factor $\gamma$. The objective is to find the policy $\pi(a | s)$ that maximises the expected return $\E_\pi[\sum_{t=1}^{T} \gamma^t r(s_t, a_t)]$. We can extend this to a goal-conditional formulation \cite{Kaelbling93learningto} to learn a policy $\pi(a | s, g)$ that maximises the expected return $r(s_t, a_t, g)$ under a goal distribution $g \sim \mathcal{G}$ as $\E_\pi[\sum_{t=1}^{T} \gamma^t r(s_t, a_t, g)]$.


At every step, the agent observes the current state $s_t$ consisting of \blue{the current arm joint configuration as well as the current end-effector pose, the next generated gripper velocities and the end-effector goal $g$, all in the robot's base frame. Rotations and changes in rotation are represented by normalised quaternions, resulting in a state space of dimension $21 + n_{joints}$ where $n_{joints}$ is the number of articulated joints in the robot's arm.} It then learns a policy $\pi(s_t, g)$ for the next base velocity commands $\mathbf{v_b}$.
We also experimented with learning the parameters of a geometric modulation as introduced in \cite{twelsche18iros} or velocities relative to the end-effector velocity but found directly learning the base commands to be more robust.

\subsection{Kinematic Feasibility}

By separating EE and base motions, we can now directly leverage the kinematical feasibility of the EE-motions to train the base policy. We convert this into a straightforward reward function that simply penalises whenever an EE-motion is not feasible. This provides us with a dense reward signal without having to rely on ground truth distances or extensive reward shaping. Instead it naturally arises from framing mobile manipulation as a modular problem. We also add a regularisation term to keep the actions small whenever possible to \sout{ensure smooth and economical behaviour}\blue{avoid unnecessary or extensive base movements}. The overall reward function can be expressed as
\begin{equation}\label{eq:objective}
    r(s, a, g) = -\mathds{1}_{!kin} - \lambda ||\mathbf{a}||^2,
\end{equation}
where $\mathds{1}_{!kin}$ is an indicator function evaluating to one when the next gripper pose is not kinematically feasible and $\lambda$ is a hyperparameter weighting the squared norm of the actions. To evaluate the kinematic feasibility, we first compute the next desired end-effector pose from the current pose and velocities. We then use standard kinematic solvers to evaluate whether this new pose is feasible.


We optimise this objective with recent model-free RL algorithms that have shown to be robust to noise and overestimation, namely TD3~\cite{fujimoto18a} and SAC~\cite{haarnoja18b}. An episode ends when either the end-effector pose is within 2.5cm of the goal or \blue{more than 19} kinematic failures have occurred \blue{(99 during evaluation)}.
The result is a reward function that can be evaluated without any adaptation across a wide variety of platforms. It furthermore allows us to use the same learned behaviours across a wide range of tasks as the RL objective is agnostic to the nature of the task itself. In our experiments, we show that EE-motions for many common robotic tasks can be easily derived from existing motion systems or constructed with very simple methods -- as we can now abstract from the feasibility of the motion.

\subsection{Training Task}

The motivation for modularising end-effector and base control is to learn a base policy that enables a large number of task-specific end-effector trajectories. In many cases we will not know all tasks at training time. To be able to generalise to diverse end-effector motions the agent should ideally observe a wide range of different relative EE-poses, motions and goals. 
To do so, we train the agent on a random goal reaching task. We first initialize the robot in a random joint configuration and then uniformly sample end-effector goals within a distance of one to five meters around the robot base and from the full range of reachable heights.

To generate end-effector motions, we use a \textit{linear dynamic system} where the end-effector velocity for each step is generated as the difference between the current pose and the sampled goal, constrained by a minimum and a maximum velocity. \green{For the orientation part we use spherical linear interpolation (slerp)}. By training with a very simple EE-\sout{planner}\blue{motion generator} that does not take into account the current joint configuration, we shift the burden of generating feasible kinematic movements to the base policy. During training we add a small Gaussian noise with a standard deviation of $\SI{1.5}{\centi\meter/s}$ to base velocities to increase robustness to imperfect motion executions in the real world.



\section{Experimental Evaluation}\label{sec:experiments}
We evaluate our approach on multiple mobile robot platforms in a series of analytical, simulated and real-world experiments to address the following questions:
\begin{itemize}
    \item Does our approach generalise across robotic platforms with different kinematic abilities?
    \item Do the learned policies generalise to task-specific gripper motions from both seen and unseen EE-motion \sout{planners}\blue{generators}?
    \item Do the analytically learned policies transfer to execution in simulation and the real world?
\end{itemize}

\subsection{Experimental setup}
\subsubsection{Robot Platforms}
\label{sec:platforms}

We train agents \green{for} three different robotic platforms differing considerably in their kinematic structure and base motion abilities. The \textit{PR2} robot is equipped with a $7$-DOF arm mounted on an omnidirectional base, giving it high mobility and kinematic flexibility. The \textit{TIAGo} robot is also equipped with a $7$-DOF arm and we additionally use the height adjustment of its torso. For the base motion it uses a differential drive restricting its mobility compared to the PR2. The \textit{Toyota HSR} robot also has an omnidirectional base but the arm is limited to $5$-DOF including the height adjustable torso. Given the low flexibility of the HSR arm, we consider distances of up to $\SI{10}{\centi\meter}$ and angles of up to $12$ degree to the desired EE-poses as kinematically feasible. This leeway does not apply to the final goal. To minimize the use of this leeway, we additionally penalize the sum of the squared distance and angular distance to the desired EE-pose, scaled into the range of $[-1, 0]$ (i.e. smaller or equal to the penalty for kinematically infeasible poses). The action space for these platforms is continuous, consisting of either one (diff-drive) or two (omni) directional velocities $\mathbf{v_{b, \{x, y\}}}$ and an angular velocity $\mathbf{v_{b, \theta}}$. \tabref{tab:constraints} shows the constraints we set across the different platforms in the analytical environment.

\setlength{\tabcolsep}{4pt}
\begin{table}
    \centering
    \begin{tabularx}{0.485\textwidth}{l|YYYY}
      \toprule
        Parameter & EE-\sout{Planner}\blue{Motion} & PR2 & TIAGo & HSR \\
      \midrule
        Max. velocity (m/s) & 0.1 & 0.2 & 0.2 & 0.2 \\
        Max. rotation (rad/s) & 0.1 & 1.0 & 0.4 & 1.5 \\
        Goal height (m) & - & [0.2, 1.4] & [0.2, 1.5] & [0.2, 1.4]\\
        Restr. height (m) & - & [0.4, 1.0] & [0.4, 1.1] & [0.4, 1.1]\\
      \bottomrule
    \end{tabularx}
    \caption{Velocity constraints for the different robot platforms and components in the analytical environment. Constraints in the physics simulator are defined by the respective default trajectory controllers. Height constraints refer to the \textit{ggr} and \textit{ggr restr} tasks and are defined for the wrist link of the robot.}
    \label{tab:constraints}
\end{table}
\setlength{\tabcolsep}{6pt}

\subsubsection{Tasks}

\begin{figure}
	\centering
	\resizebox{\columnwidth}{!}{%
  	\begin{tabular}{cc}
  		\includegraphics[width=0.31\columnwidth,trim={11cm 1cm 11cm 1cm},clip,angle =0]{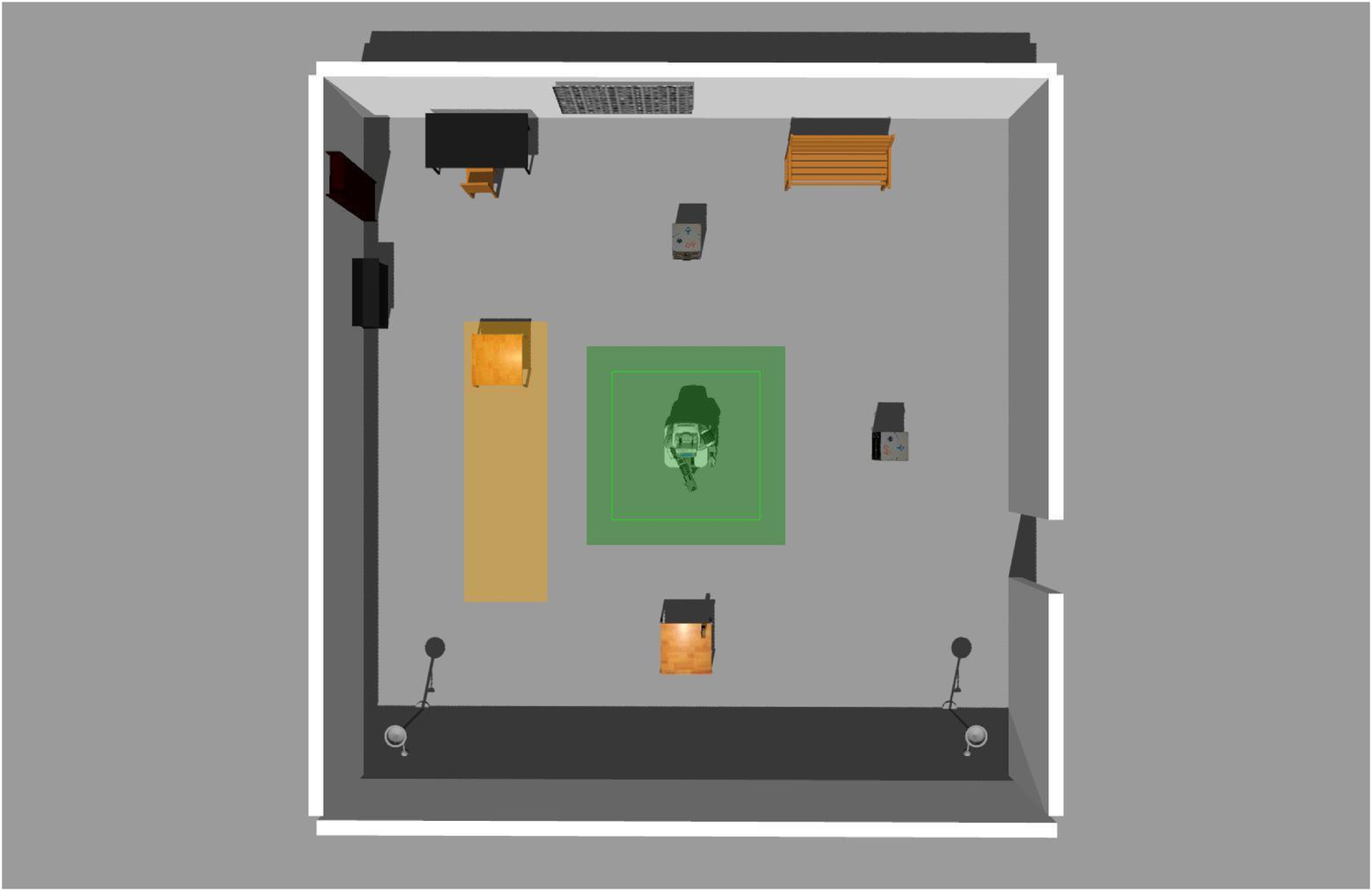} &
  		\includegraphics[width=0.31\columnwidth,trim={7.0cm 1.0cm 7.0cm 1.0cm},clip,angle =0]{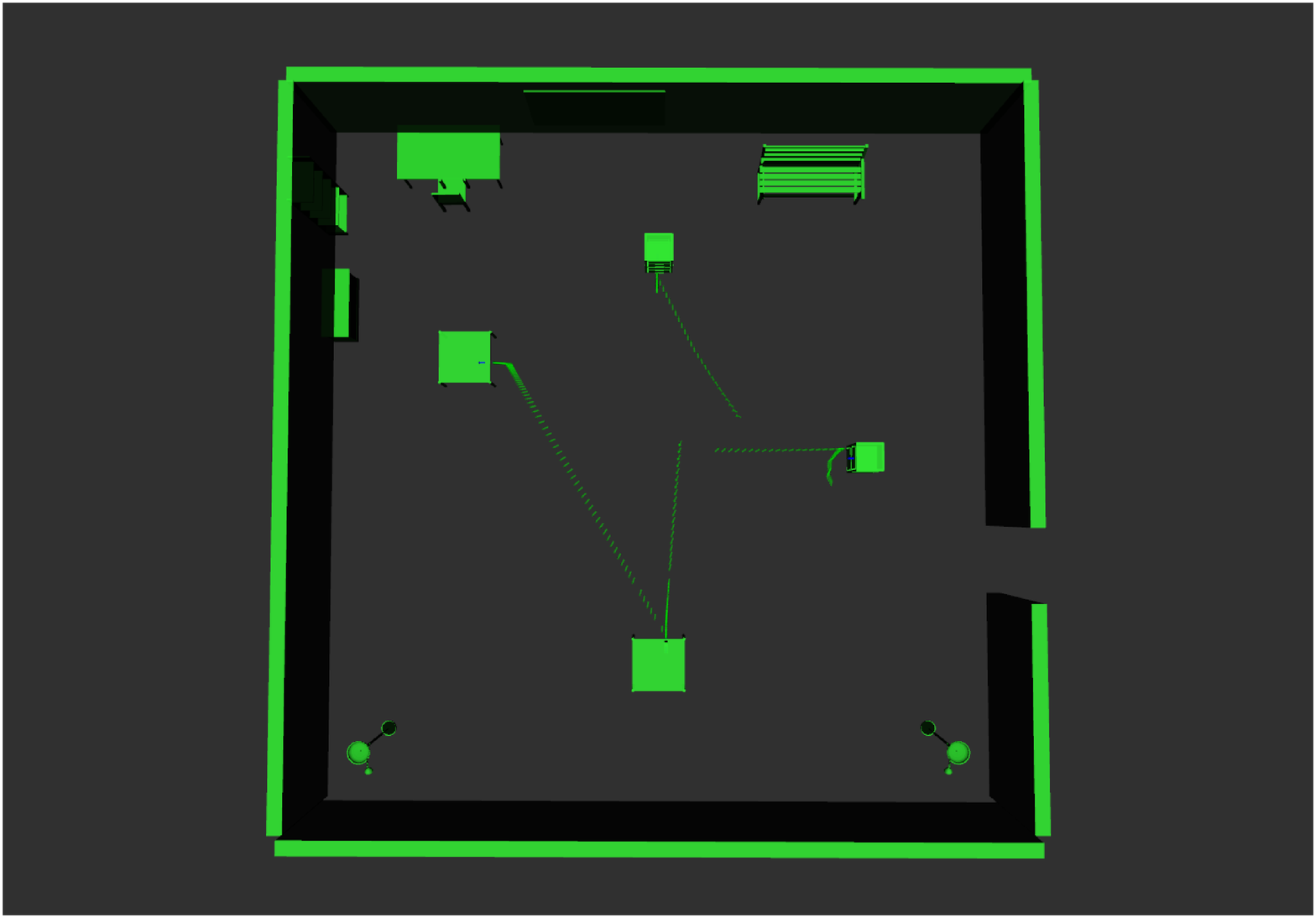}
	\end{tabular}
	}
    \caption{Left: The robot starts in a random location and rotation in the center of the room (green). The task objects consisting of tables and shelves equipped with doors and drawers are located around the robot. The random location of the drop-off table is marked in orange. Right: Examples of the different generated EE-motions for each task. Markers indicate the desired EE-pose at every $10^{th}$ step.}
    \label{fig:modulation_tasks}
\end{figure}

We construct five tasks: A \textit{general goal reaching (ggr)} task in which the goals are selected randomly as in the training phase. As the kinematics become very restrictive towards the edges of the height range, we also analyse the results for initial configurations and goals that are restricted to more common heights, which we refer to as \textit{ggr restr}. \tabref{tab:constraints} shows the values that we specify for the different robots. A \textit{pick\&place} task in which the robot has to grasp an object randomly located on the edge of a table, move it to a different goal table randomly located on another wall in the room and place it down. We use the linear system to generate EE-motions by sequentially combining four goals relative to the object: slightly in front of the object, at the object, in front of and at the goal location. To test whether our approach generalises to different EE-motions we then construct two more tasks from an unseen \textit{imitation learning system} developed in~\cite{twelsche17iros}. These motions are learned from a human teacher and encoded in a dynamic system following a demonstrated hand trajectory to manipulate a certain object. As a result, the motions can differ substantially from the linear system used during training. We use models to \textit{grasp and open a cabinet door} as well as to \textit{grasp and open a drawer}. The corresponding motion models are autonomously adapted to the given poses of the handled objects. \figref{fig:modulation_tasks}~(right) shows examples of the generated EE-motions for each of the tasks.

For the virtual evaluation we locate the objects in a room around the robot as shown in \figref{fig:modulation_tasks}~(left). In all tasks, the robot starts randomly within a $1.5\times \SI{1.5}{\meter}$ square in the center of the room, rotated between $[-\pi/2, \pi/2] \,\SI{}{\rad}$ relative to the first end-effector goal and in a random joint configuration sampled from the full possible configuration space, including difficult and unusual poses.

\setlength{\tabcolsep}{4pt}
\renewcommand{\arraystretch}{0}
\setlength{\tabcolsep}{6pt}
\renewcommand{\arraystretch}{1}

\subsubsection{Baselines}

We construct separate baselines for the linear system and the imitation system. For the linear system the baseline replicates the same robot base motion in $xy-$direction as the end-effector. For Tiago this agent also takes into account the limitations of the differential drive. This simple strategy removes some of the main difficulties by keeping a fixed distance between EE and body, avoiding situations in which the gripper would have to pass around or through the robot body. The baseline for the imitation system follows base motions that were learned together with the end-effector, adapted to the different lengths of the robot arms. We combine these two baselines under the terms \textit{PR2\_bl, Tiago\_bl} and \textit{HSR\_bl}. We also compare against the geometric modulation for the PR2 presented in \cite{twelsche18iros}, termed \textit{PR2\_gm}. This approach learns an approximation to the inverse reachability in closed form and modulates the base velocities to stay within allowed poses. As the approximations rely on robot specific knowledge, it cannot easily be adapted to the other platforms.

\subsubsection{Training Details and Metrics}

For each robotic platform, we conduct a hyperparameter search over the parameters listed in \tabref{tab:hyperparams}. We then select the best configuration and train the agent on new seeds for roughly 2,000 episodes (1,000,000~steps) of the random goal reaching task described in \secref{sec:approach}. For each platform we train models on ten different seeds and average the results. We then evaluate the agent's ability to achieve the described tasks without any task-specific fine-tuning over 50 episodes per task. For each task, we report the share of the trajectories executed without a single kinematic failure. Note that this is a fairly strict metric and in many cases even with a few failures, the task can still be completed successfully. For this reason, we also report the share of episodes which never deviate more than $\SI{5}{\centi\meter}$ from the EE-motions (in brackets). 
We find that both TD3 and SAC learn to solve the tasks successfully, but SAC generally results in more robust policies and ultimately slightly better performance. \blue{Therefore we use SAC throughout the remainder of this work.}

\setlength{\tabcolsep}{2pt}
\begin{table}
    \centering
    \begin{tabularx}{0.485\textwidth}{lc|lY}
      \toprule
        Parameter & Values & Parameter & Values \\
      \midrule
        Algorithm   & \{SAC, TD3\} & $\tau$  & \{0.001, 0.005\}\\
        Batch size  & \{64, 256\} & $\gamma$   & \{0.98, 0.99, 0.999\}\\
        Ik fail thresh & \{1, 19, 99\} & $\epsilon$-noise   & \{0.25, 0.5, 0.75\}\\
        $\lambda$   & \{0.0, 0.01, 0.1\} & Rnd steps   & \{0, 50'000\}\\
        Lr   & \{$\expnumber{3}{-4}$, $\expnumber{1}{-4}$, $\expnumber{1}{-5}$ \} & Policy noise   & \{0.1, 0.25, 0.5\}\\
        Lr decay  & \{0.999\} & Entropy reg & \{learn, 0.1, 0.2, 0.3\}\\
        Buffer size   & \{100'000\} & \\
      \bottomrule
    \end{tabularx}
    \caption{Hyperparameters searched, \blue{values chosen based on grid search on the training task}. We use a public implementation of the TD3 and SAC algorithms~\cite{stable-baselines3}. Parameters not mentioned are left at their default values, including the actor and critic networks of two fully-connected layers size (256, 256) for SAC and (400, 300) for TD3. IK fail thresh is the maximum number of failures before we terminate the episode and Lr is the learning rate.}
    \label{tab:hyperparams}.
\end{table}
\setlength{\tabcolsep}{6pt}



\begin{figure}
\centering
\includegraphics[width=0.485\textwidth,trim={0.0cm 0.0cm 1.0cm 1.5cm},clip,angle =0]{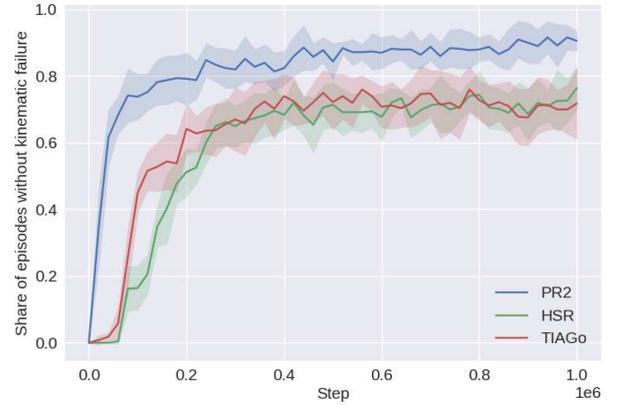}
\caption{Training progress measured as share of episodes with zero kinematic failures on the random goal reaching task. \blue{All agents were trained with SAC,} evaluated over 50 episodes and averaged over 10 seeds. Shaded regions show the standard deviation.}
\label{fig:train_curves}
\end{figure}

\subsection{Analytical Evaluation}

We train the model on analytically generated trajectories, i.e. we integrate the system step-by-step to generate the poses of robot end-effector and base. At each step we evaluate the kinematic feasibility of the generated poses without a physical simulation of robot controllers. We evaluate this system at a step size corresponding to a frequency of $\SI{10}{\hertz}$.
\figref{fig:train_curves} shows the success rates over the course of the training, averaged over ten seeds for each of the platforms. As the kinematic feasibility provides us with a dense reward, the agents reach reasonable performance already within a few hundred episodes (an average training episode lasts roughly 500 steps).

We then evaluate the trained models across all tasks in the same environment. The results are summarised in \tabref{tab:analytical}. On the PR2 the baseline that replicates the EE-velocities is able to complete between 60\% and 73\% of the linear motion tasks successfully. By simply "sliding" towards the goal, the manipulation task is kept relatively static and most of the difficulty is transferred to the IK solver. As a consequence, performance on more constraint platforms drops to between 0\% and 24\% for both the 5-DOF arm of the HSR as well as the diff-drive of TIAGo that has to "circle" towards the goal. This illustrates the need for base and arm to work together to achieve these motions. On the door and drawer tasks, the PR2 and TIAGo are able to follow some of the imitated motions, but fail in the majority of cases with between 9\% and 32\% of the episodes fully successful. The HSR is completely unable to follow these motions.

The geometric modulation approach \textit{PR2\_gm} is not able to significantly improve further upon the baseline, indicating that the velocities of the baseline that serve as input are already removing most of the difficulty from the task. Failure cases for the geometric modulation include when the starting pose lies outside the approximated reachability range as well as situations in which the EE comes close to the edges of the approximation.


In contrast, our approach to directly learn the base velocities and kinematic feasibility achieves high success rates across all robots and tasks, solving the goal reaching task with 90.0\% for PR2, 71.6\%  for TIAGo, and 75.2\% for HSR.
This translates into near perfect performance on the pick\&place task with 97.0\% success for PR2, 91.4\% for TIAGo and 90.2\% for HSR.
Looking at the imitation tasks, we find that these results also generalise to the unseen motions with all platforms achieving 90.6\% or more of all episodes without a single kinematic failure. This indicates a number of things. First, our training task is comparably difficult, making it a good training ground to train for general tasks. This is expected as the goal distribution encompasses the full range of possible heights in which the kinematics can become quite restrictive. Focusing on a workspace restricted to common heights (\textit{ggr restr}), performance increases for both TIAGo (+8.8\%) and HSR (+4.9\%) while remaining relatively unchanged for the PR2 (-1.4\%). Secondly, we find no drop in performance while following the unseen imitation learning system which demonstrates the agent's ability to enable general movements of the end-effector.

We observe that the performance improves further if we consider all episodes that never deviate more than $\SI{5}{\centi\meter}$ from the desired motions (values in brackets) for both the PR2 and TIAGo. The performance drops slightly for the HSR as this is a tighter measure than the leeway that we grant its IK solutions as discussed in \secref{sec:platforms}. At the same time, the overwhelming majority of episodes does not use most of this leeway, with differences in the two metrics of only between 2.0\% and 7.5\% across all tasks. On average, the distance to the exact desired EE-motion is only $\SI{1.6}{\centi\meter}$ across all the successful steps.

\begin{figure}
	\centering
	\resizebox{\columnwidth}{!}{%
  	\begin{tabular}{ccc}
  		\fbox{\includegraphics[width=0.32\columnwidth,trim={0.0cm 0.0cm 0.0cm 0.0cm},clip,angle =0]{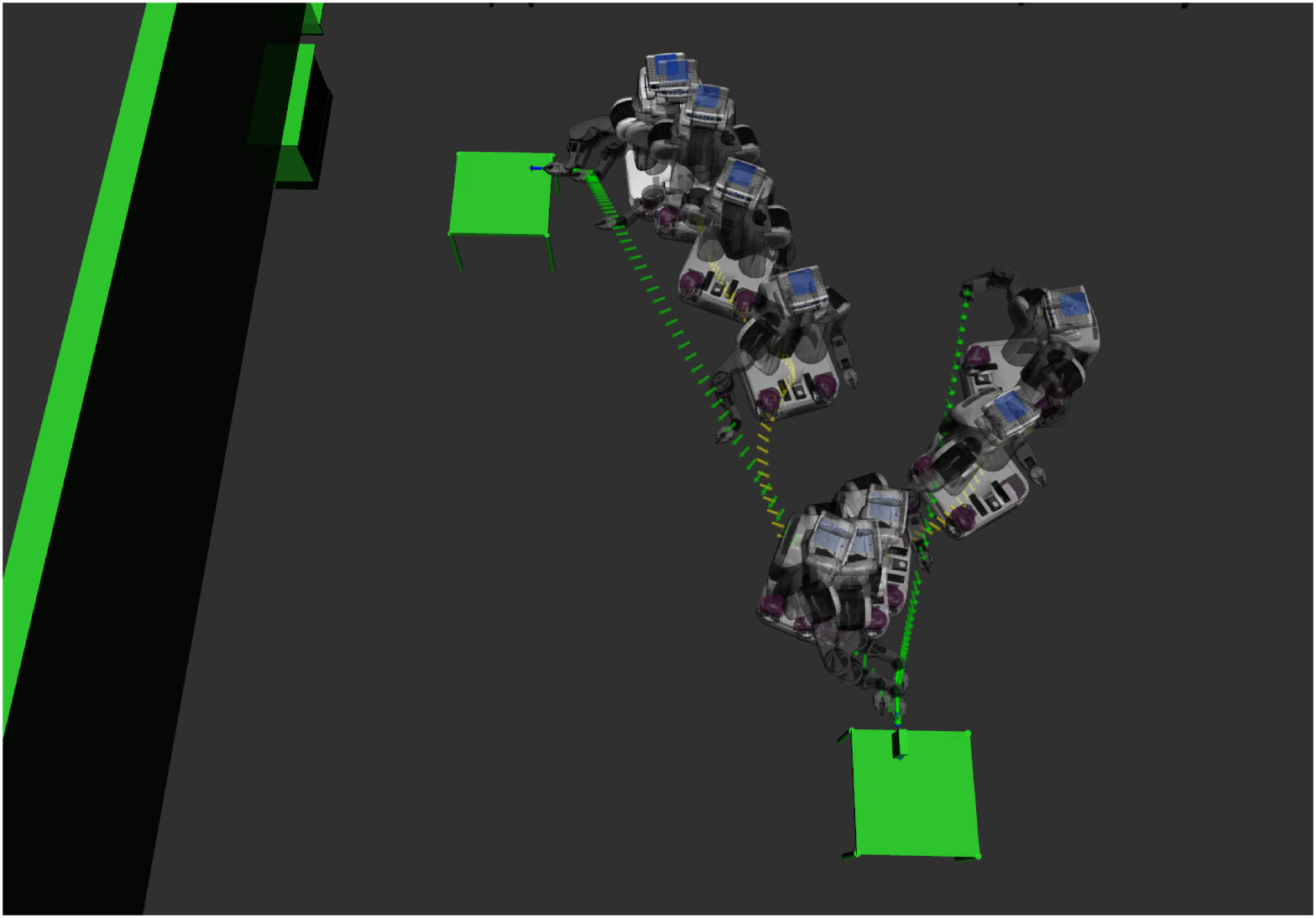}} &
  		\fbox{\includegraphics[width=0.32\columnwidth,trim={0.0cm 0.0cm 0.0cm 0.0cm},clip,angle =0]{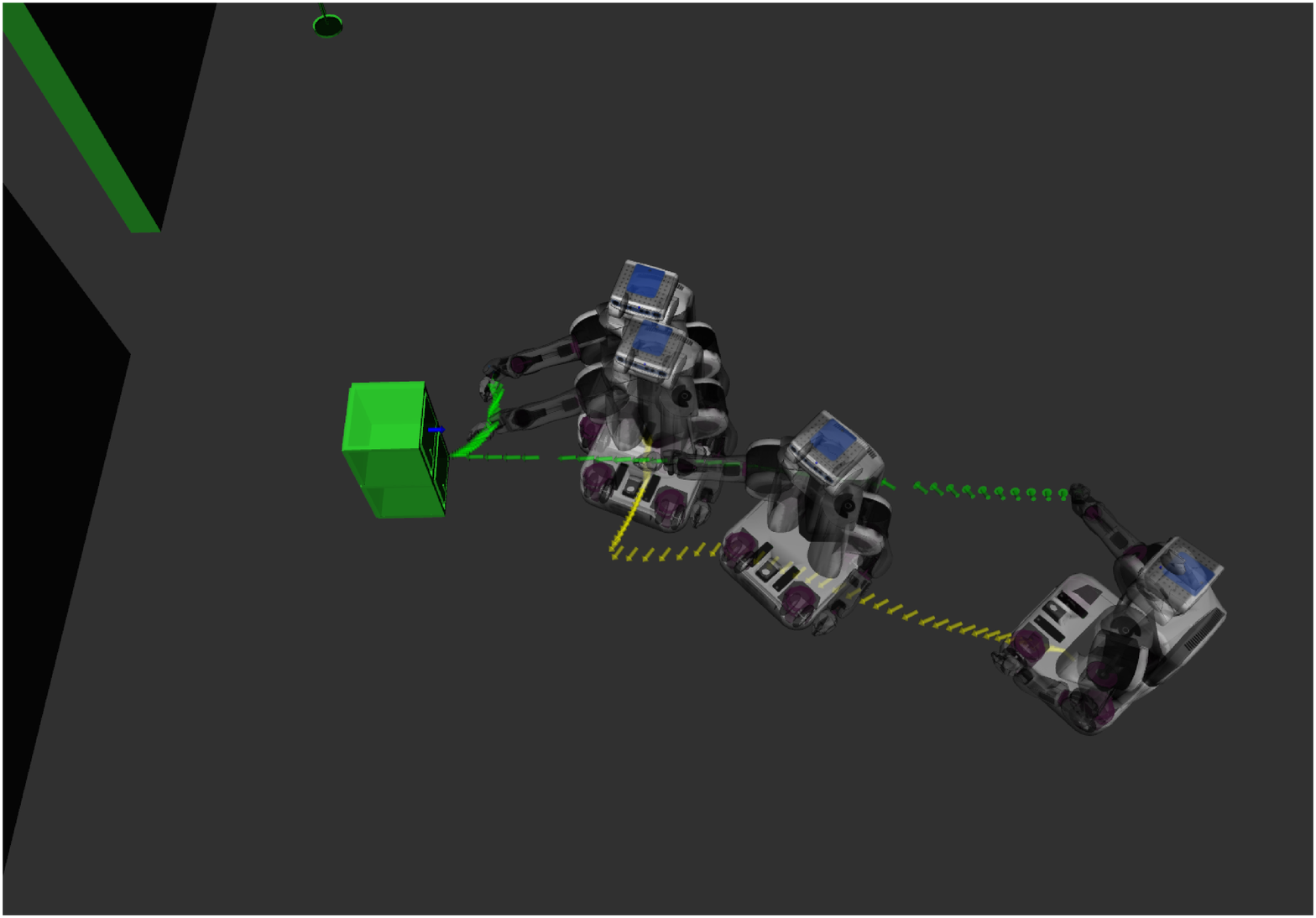}} &
  		\fbox{\includegraphics[width=0.32\columnwidth,trim={0.0cm 0.0cm 0.0cm 0.0cm},clip,angle =0]{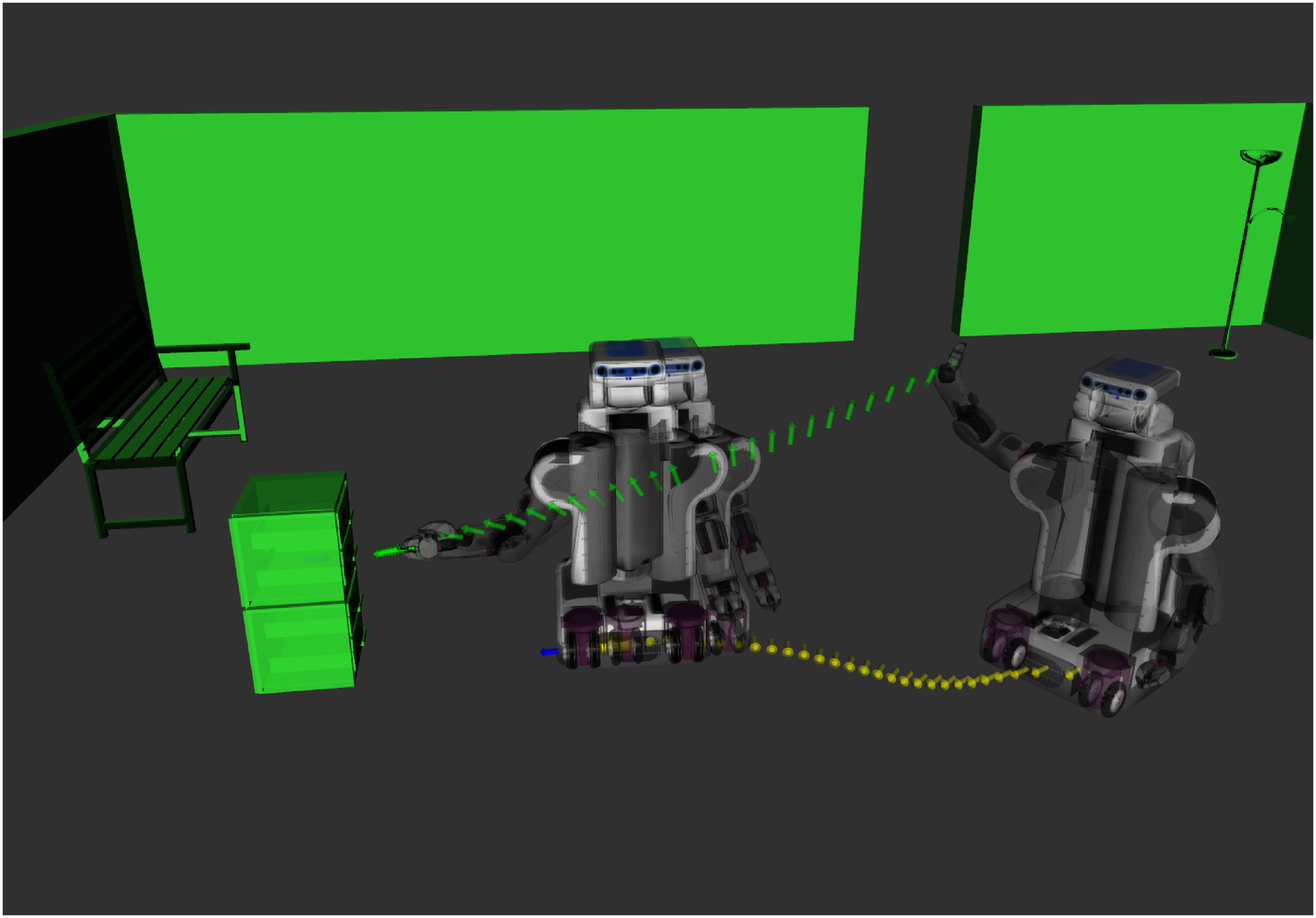}}\\  		
 		\fbox{\includegraphics[width=0.32\columnwidth,trim={0.0cm 0.0cm 0.0cm 0.0cm},clip,angle =0]{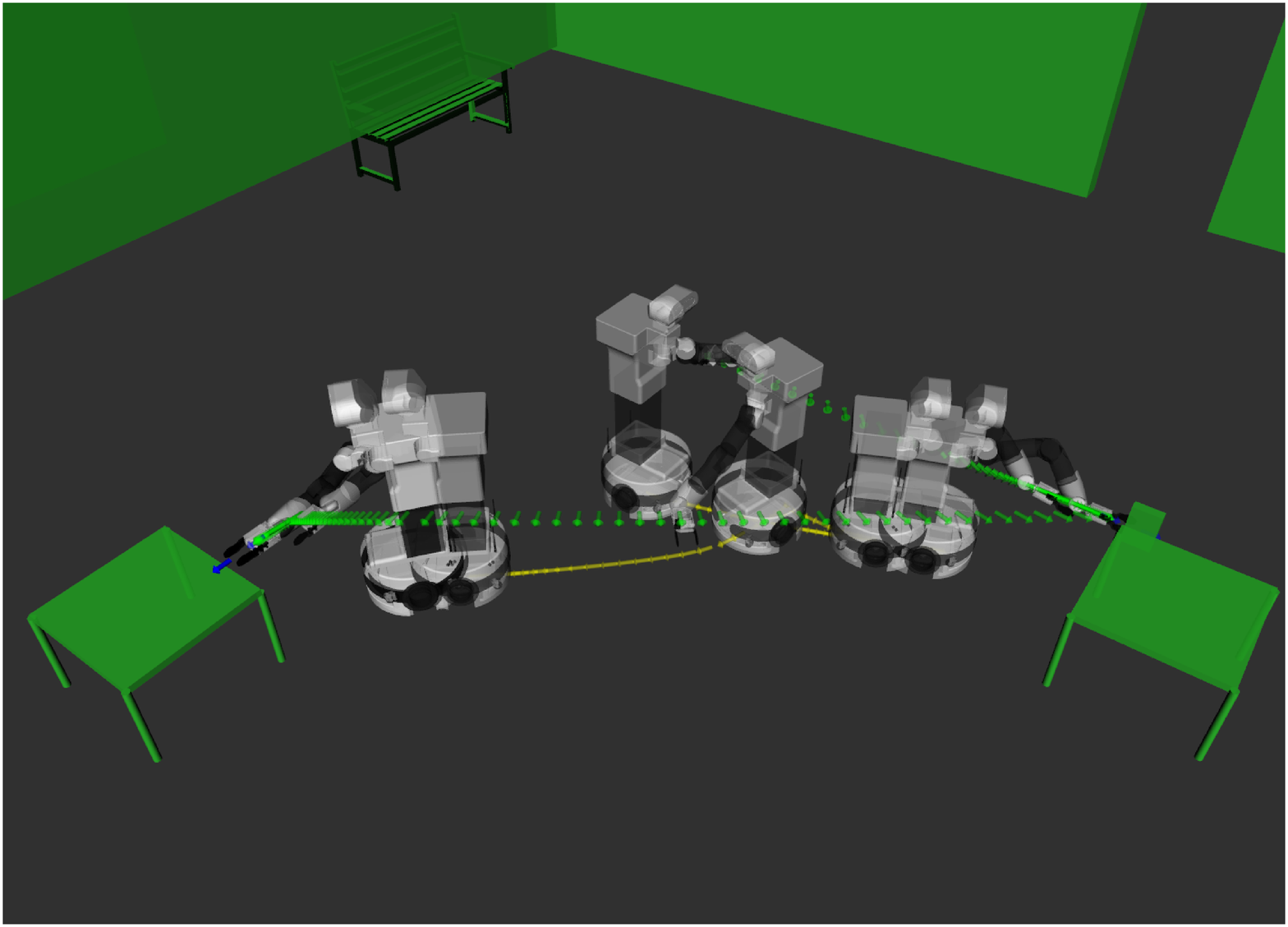}} &
 		\fbox{\includegraphics[width=0.32\columnwidth,trim={0.0cm 0.0cm 0.0cm 0.0cm},clip,angle =0]{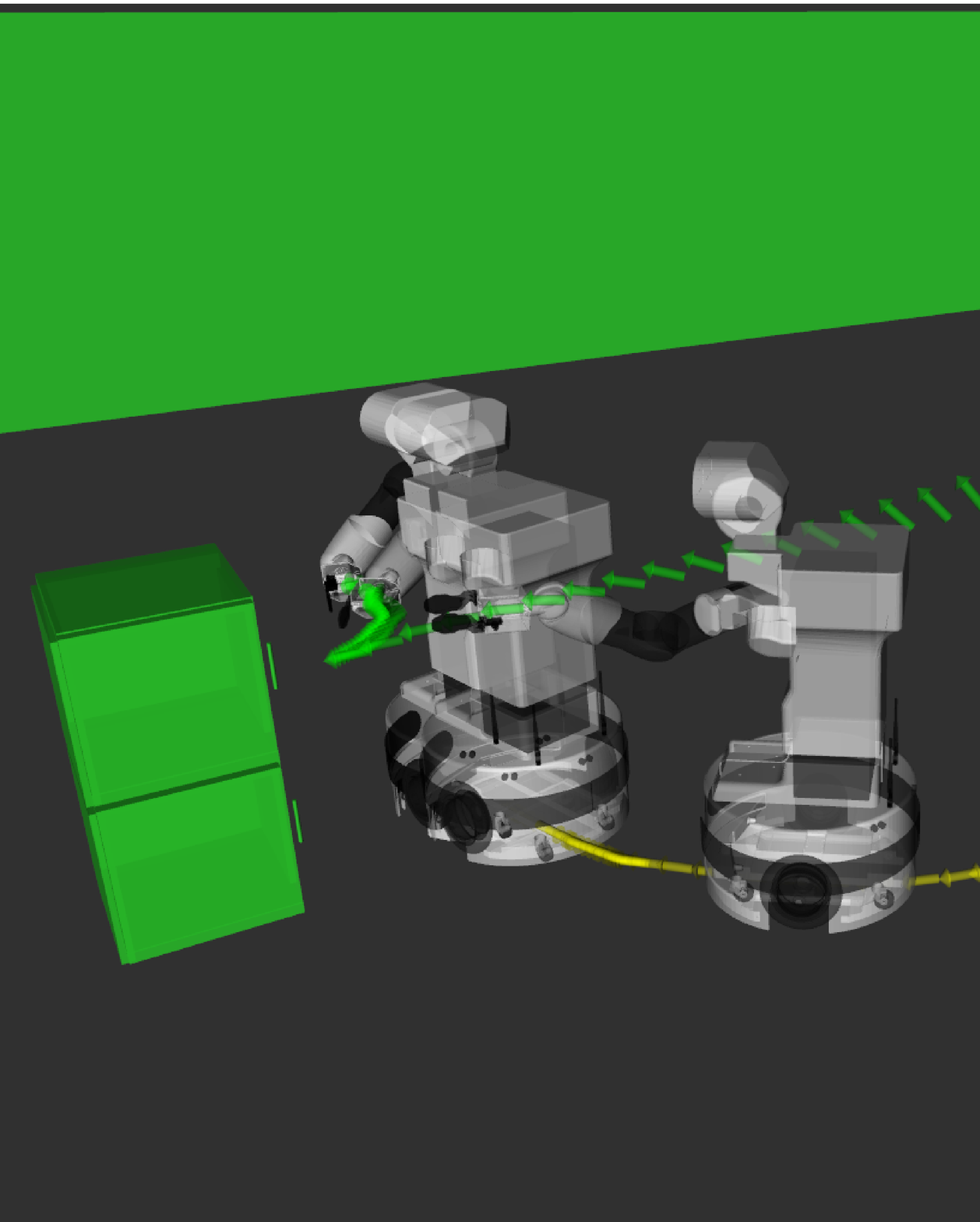}} &
 		\fbox{\includegraphics[width=0.32\columnwidth,trim={0.0cm 0.0cm 0.0cm 0.0cm},clip,angle =0]{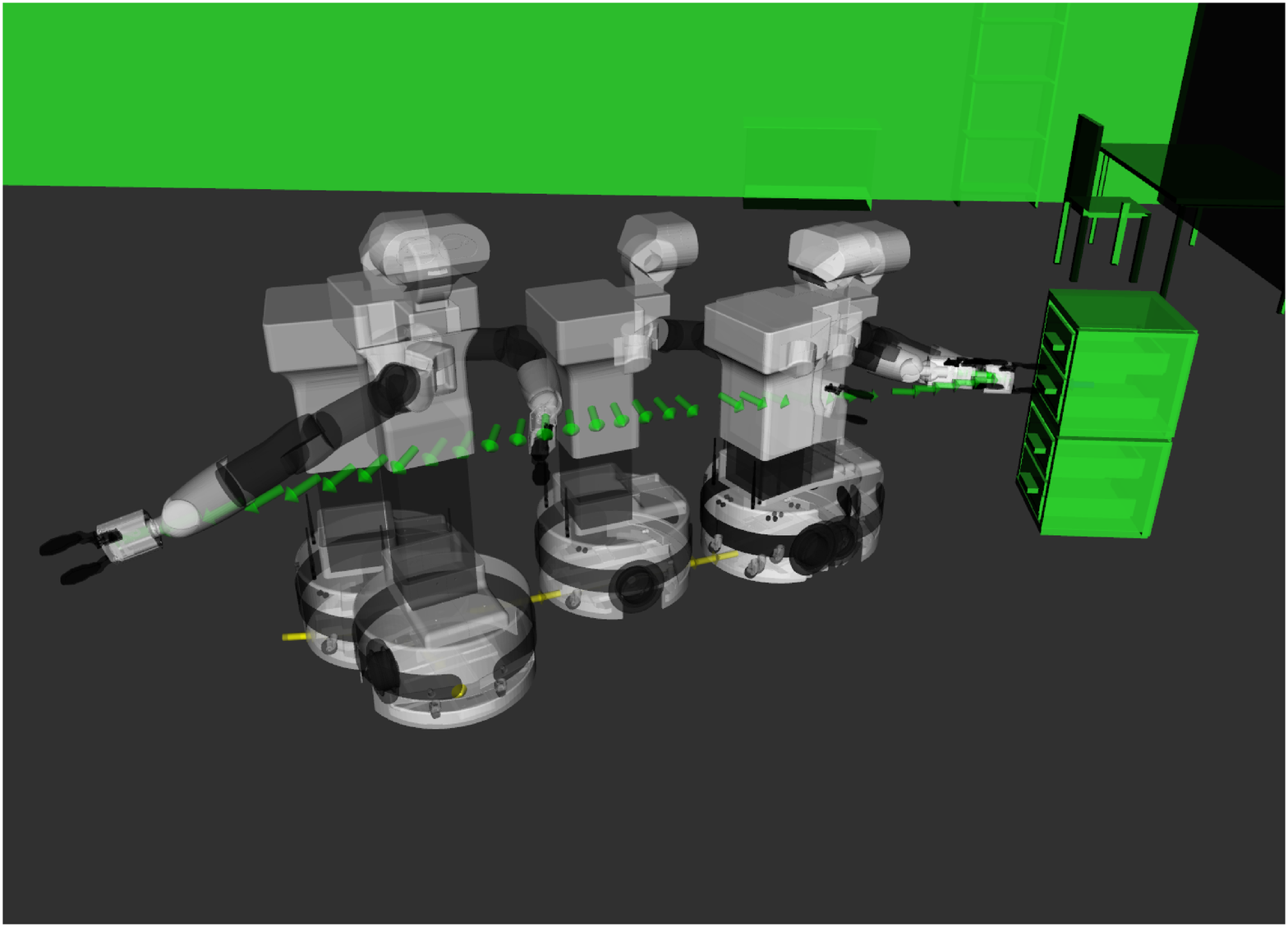}}\\
 		\fbox{\includegraphics[width=0.32\columnwidth,trim={0.0cm 0.0cm 0.0cm 0.0cm},clip,angle =0]{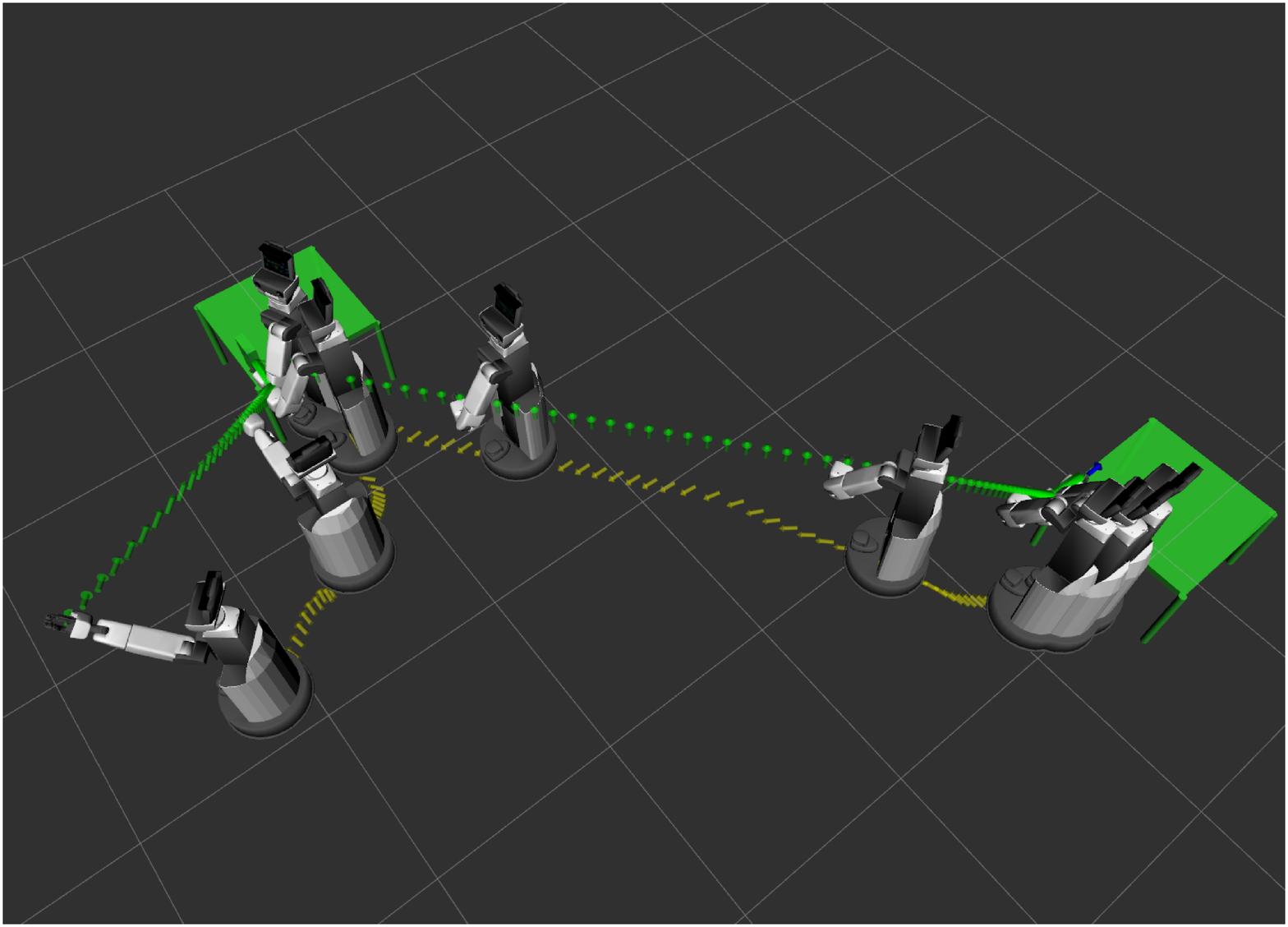}} &
 		\fbox{\includegraphics[width=0.32\columnwidth,trim={0.0cm 0.0cm 0.0cm 0.0cm},clip,angle =0]{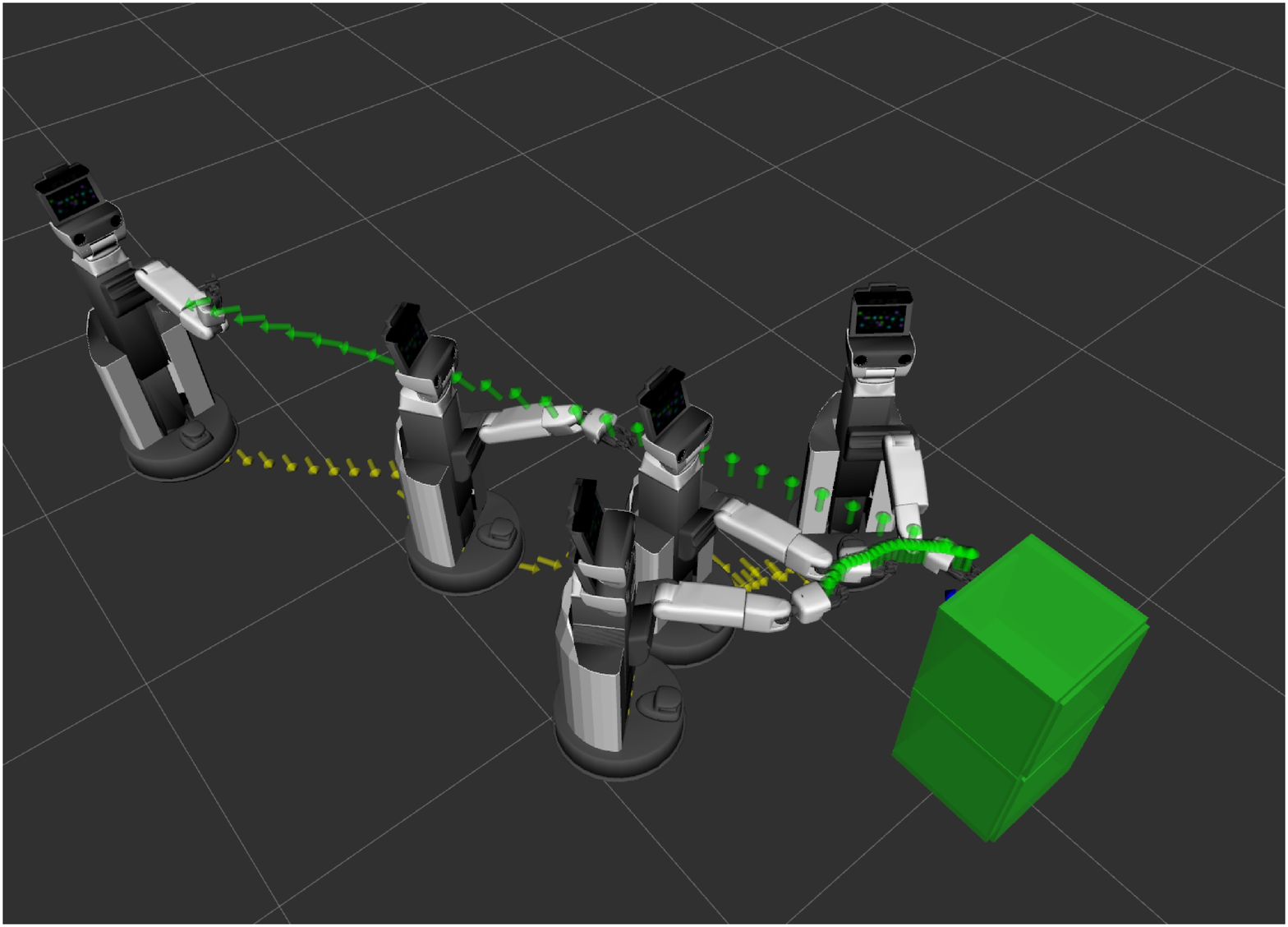}} &
 		\fbox{\includegraphics[width=0.32\columnwidth,trim={0.0cm 0.0cm 0.0cm 0.0cm},clip,angle =0]{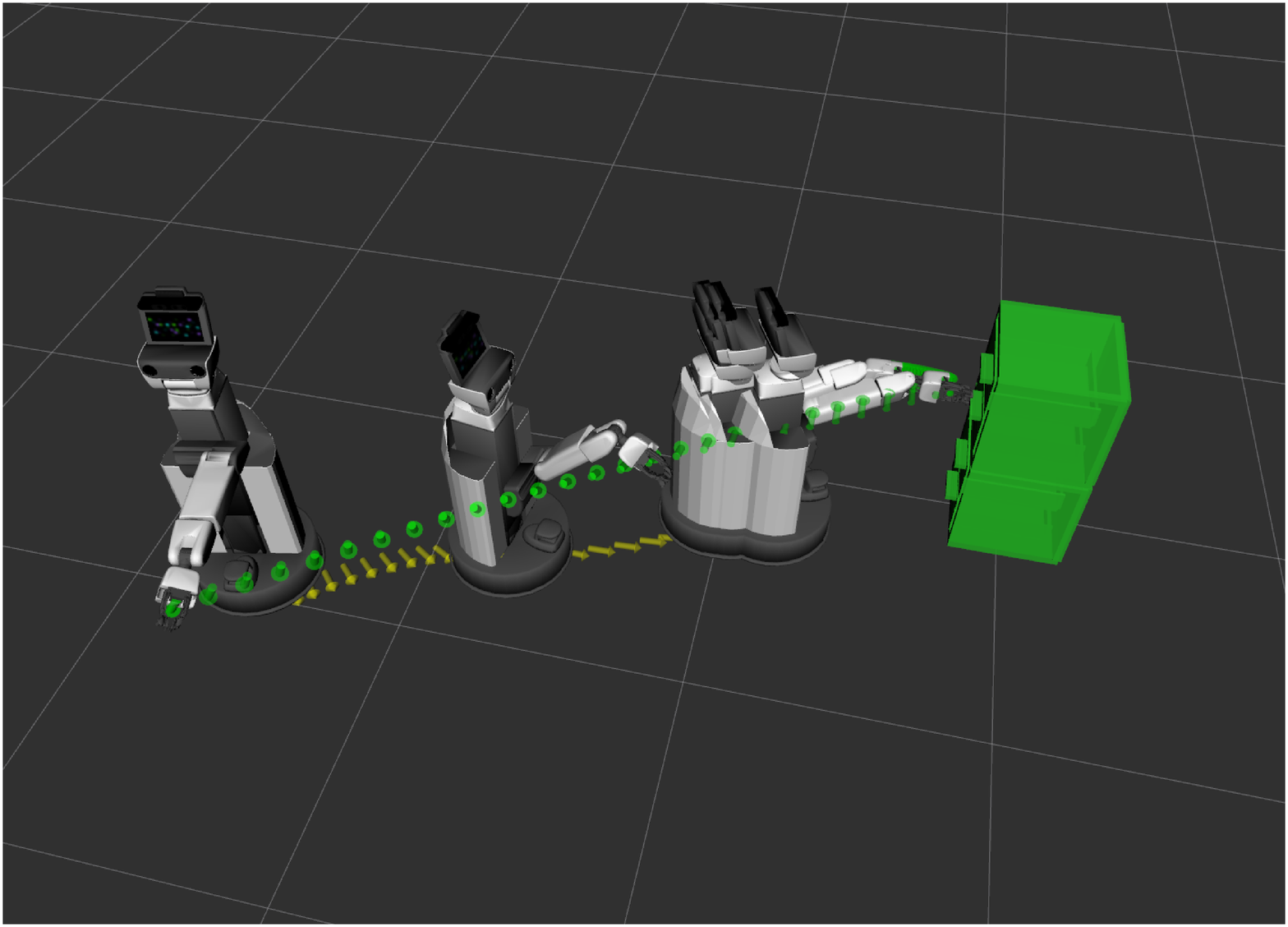}}
	\end{tabular}
	}
	\caption{Analytical evaluation on the pick\&place (left), door (mid) and drawer opening (right) tasks for the PR2 (top), TIAGo (mid) and the HSR (bottom). Markers show the base (yellow) and EE-trajectory (green).}
  	\label{fig:analytical_eval}
\end{figure}

\begin{figure}
	\centering
	\resizebox{\columnwidth}{!}{%
  	\begin{tabular}{cc}
  		\includegraphics[width=0.48\columnwidth,trim={0.0cm 1.0cm 5.0cm 5.0cm},clip,angle =0]{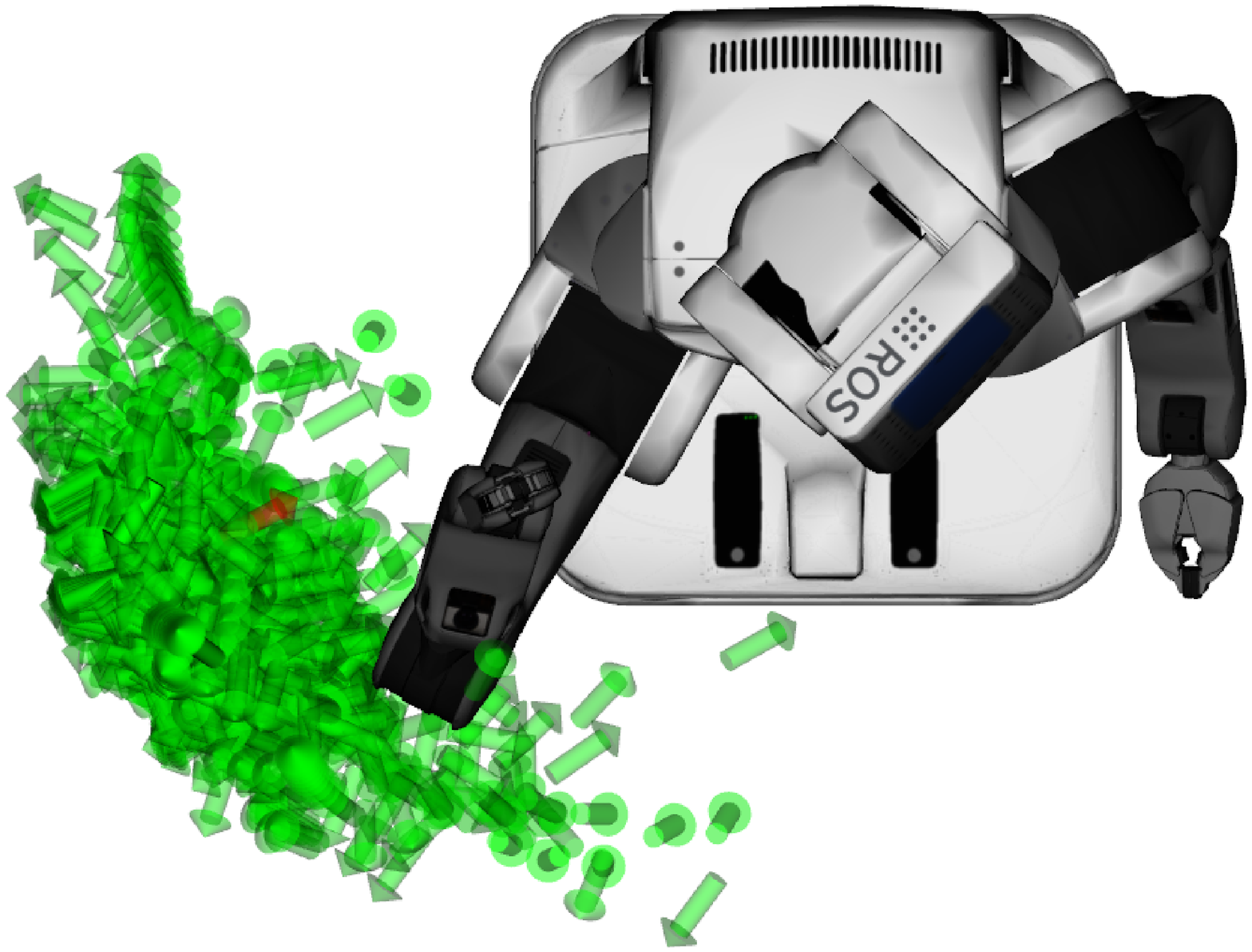} &
  		\includegraphics[width=0.48\columnwidth,trim={6.0cm 5.0cm 6.0cm 4.0cm},clip,angle =0]{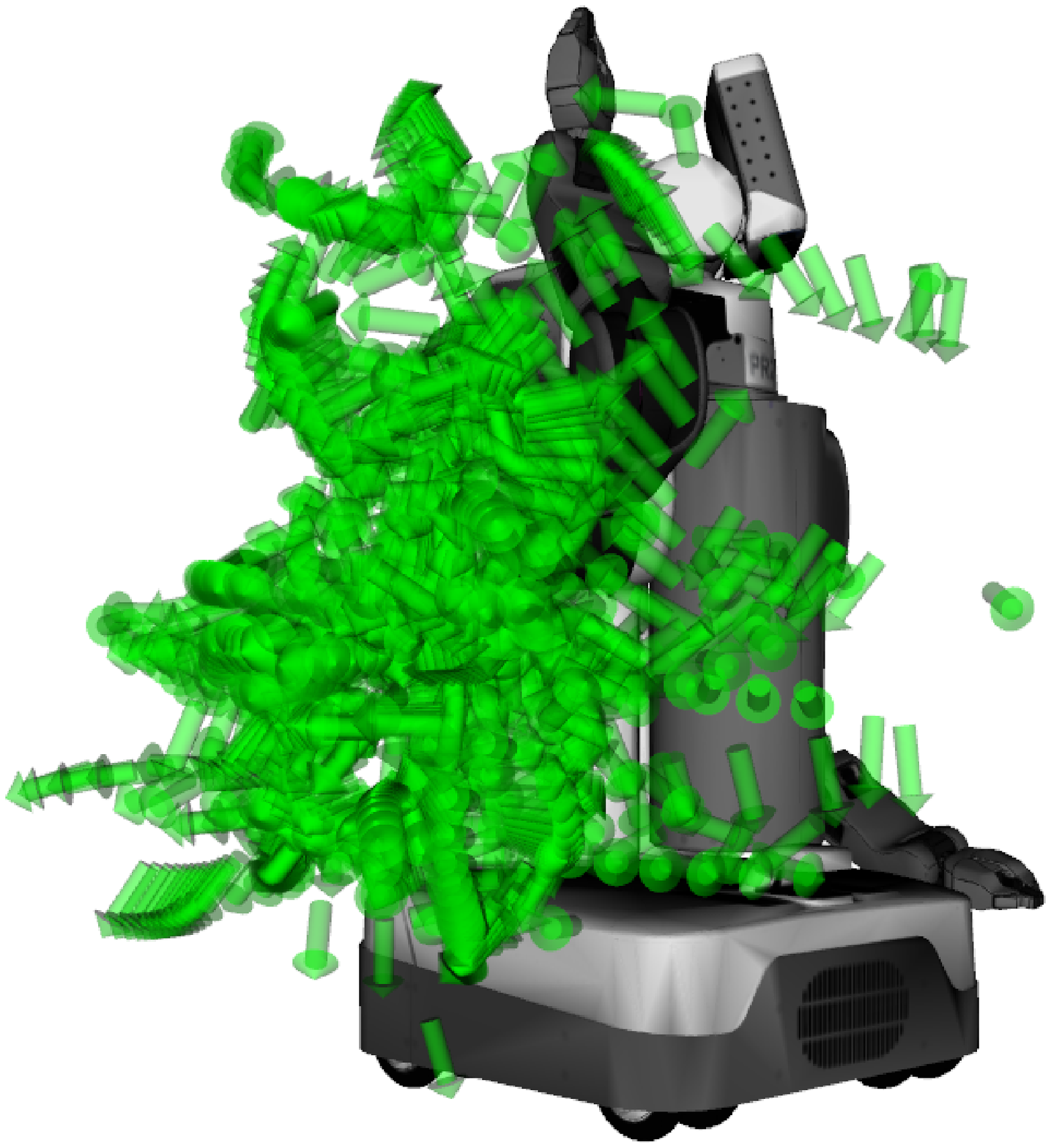}
	\end{tabular}
	}
    \caption{Covered work space area of the learned PR2 agent: relative end-effector poses encountered over 50 episodes of the \textit{ggr} task, evenly subsampled to 1000 poses. Red indicates kinematically infeasible poses.}
  	\label{fig:relpose}
\end{figure}

Qualitatively, we find that the resulting policies take reasonable and \sout{economical}\blue{practical} paths. These include behaviours such as rotating around the closest angle, seeking out robust relative EE-poses, e.g. the PR2 moving sidewards with the right arm in front towards the goal such that the arm has the most freedoms, or if the EE pose is on the opposite side of robot body the learned strategy is to first dodge sideways to bring the end-effector in front of itself. Examples are depicted in \figref{fig:analytical_eval} and in the accompanying video.
To qualitatively examine the diversity of the learned policies, in figure \ref{fig:relpose} we plot the work space area covered by the end-effector for the PR2 agent. We find that the agent does not decay to singular behaviours, but rather uses a large area of the possible EE-poses.


\setlength{\tabcolsep}{2pt}
\begin{table}
    \centering
    \begin{tabularx}{0.485\textwidth}{l|YYY|Y Z}
    \toprule
      Agent   & \multicolumn{3}{c|}{Linear Dynamic System} & \multicolumn{2}{c}{Imitation Learning}\\
     \cmidrule{2-6}
            & ggr & ggr restr & pick\&place & door & drawer \\
      \midrule
        PR2\_bl   & 60.8 (65.0) & 70.8 (72.8) & 72.6 (76.8) & 30.2 (33.6) & 31.6 (35.4)\\
        PR2\_gm   & 64.6 (69.6) & 68.6 (74.4) & 74.4 (78.6) & 31.6 (38.0) & 28.2 (34.0)\\
        \textbf{PR2} & \textbf{90.2 (91.2)} & \textbf{88.8 (90.6)} & \textbf{97.0 (97.4)} & \textbf{94.2 (95.4)} & \textbf{95.4 (95.4)}\\
        \cmidrule{1-6}
        Tiago\_bl & 21.6 (24.6) & 23.6 (26.8) & 12.0 (13.8) & \phantom{0}9.2 (10.0) & 28.6 (31.8)\\
        \textbf{Tiago} & \textbf{71.6 (73.4)} & \textbf{80.2 (83.0)} & \textbf{91.4 (92.2)} & \textbf{95.3 (96.9)}  & \textbf{94.9 (95.3)} \\
        \cmidrule{1-6}
        HSR\_bl   & \phantom{0}5.4 \phantom{0}(7.4) & \phantom{0}7.6 (10.4) & \phantom{0}0.0 \phantom{0}(3.2) & \phantom{0}0.0 \phantom{0}(0.0) & \phantom{0}0.0 \phantom{0}(0.0)\\
        \textbf{HSR} & \textbf{75.2 (69.8)} & \textbf{80.1 (72.6)} & \textbf{93.4 (90.2)} & \textbf{91.2 (87.0)} & \textbf{90.6 (88.6)}\\
        \bottomrule
    \end{tabularx}
    \caption{Performance in the analytical evaluation as share of successfully executed episodes with zero kinematically infeasible EE-poses and share of episodes that never deviate more than $\SI{5}{\centi\meter}$ from the EE-motion (in brackets). The bold rows represent our proposed approach. We evaluate each task over 50 episodes for 10 random seeds.}
    \label{tab:analytical}
\end{table}
\setlength{\tabcolsep}{6pt}

\setlength{\tabcolsep}{1pt}
\renewcommand{\arraystretch}{1}
\setlength{\tabcolsep}{6pt}
\renewcommand{\arraystretch}{1}

\setlength{\tabcolsep}{1pt}

\renewcommand{\arraystretch}{1}
\setlength{\tabcolsep}{6pt}
\renewcommand{\arraystretch}{1}

\subsection{Motion Execution in Simulation}

\begin{figure}
	\centering
	\resizebox{\columnwidth}{!}{%
  	\begin{tabular}{ccc}
  		\fbox{\includegraphics[width=0.32\columnwidth,trim={0.0cm 0.0cm 0.0cm 0.0cm},clip,angle =0]{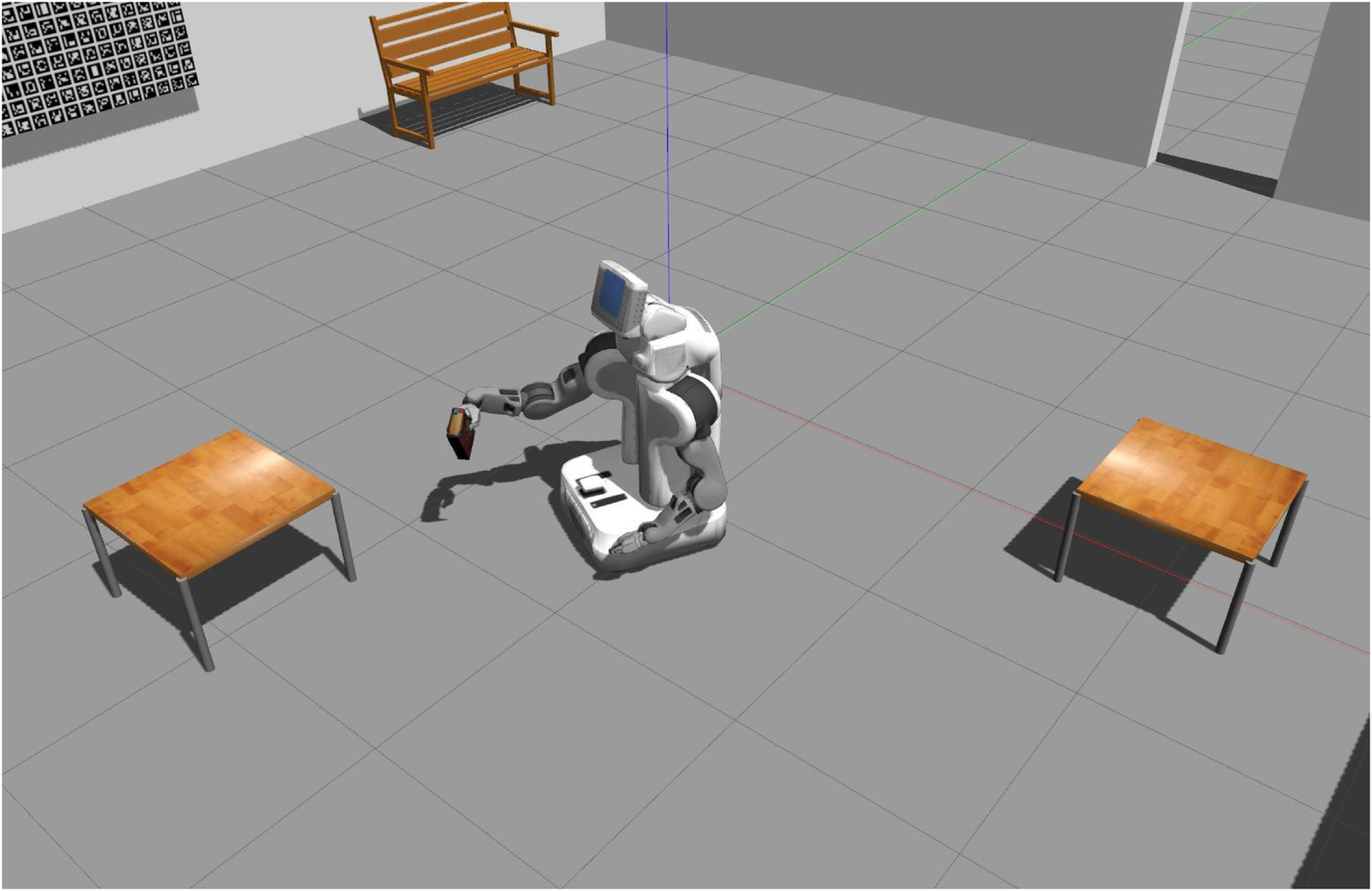}} &
  		\fbox{\includegraphics[width=0.32\columnwidth,trim={8.0cm 10.0cm 7.5cm 0.0cm},clip,angle =0]{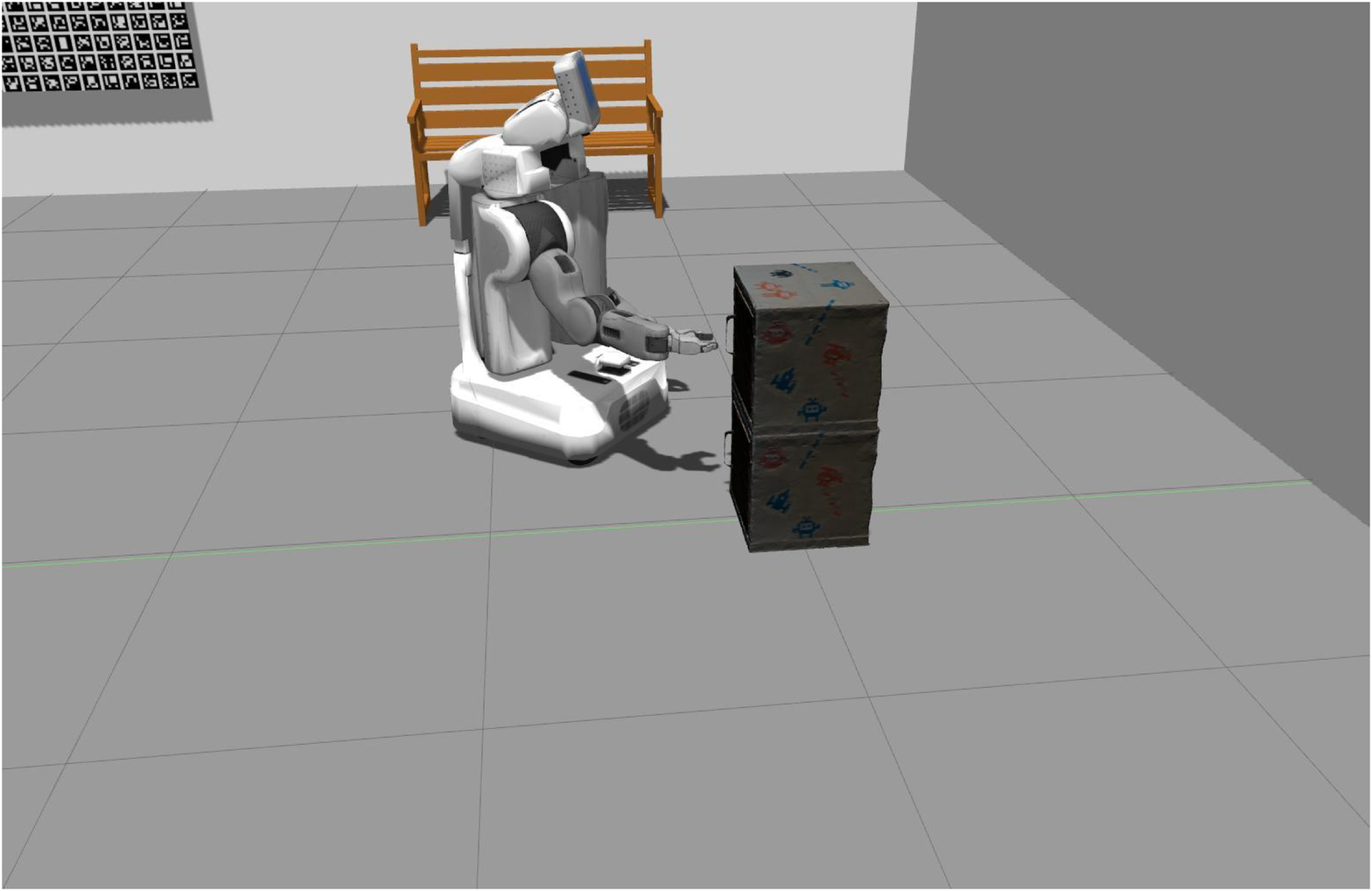}} &
  		\fbox{\includegraphics[width=0.32\columnwidth,trim={0.0cm 0.0cm 0.0cm 0.0cm},clip,angle =0]{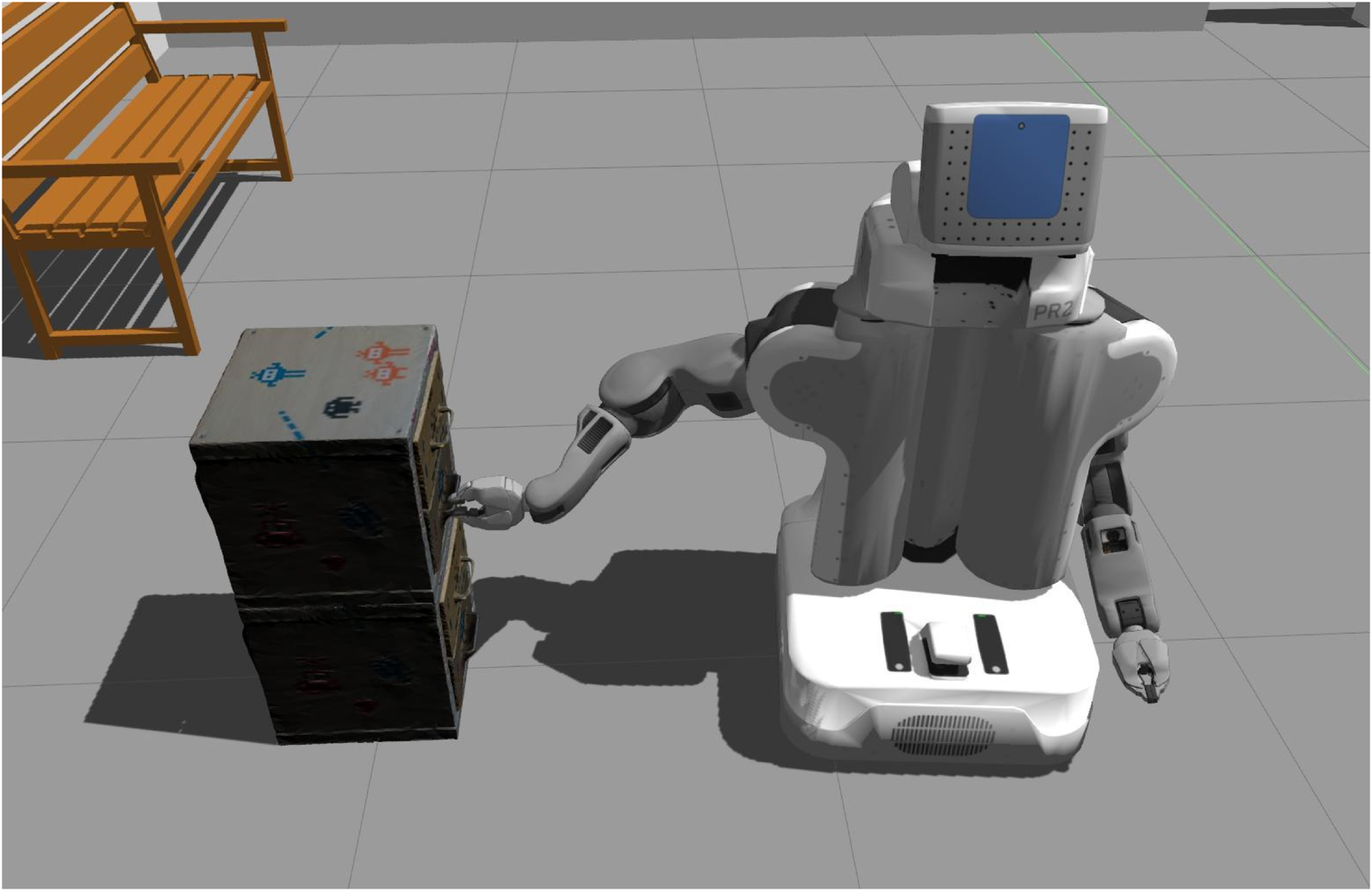}}\\  		
 		\fbox{\includegraphics[width=0.32\columnwidth,trim={0.0cm 0.0cm 0.0cm 0.0cm},clip,angle =0]{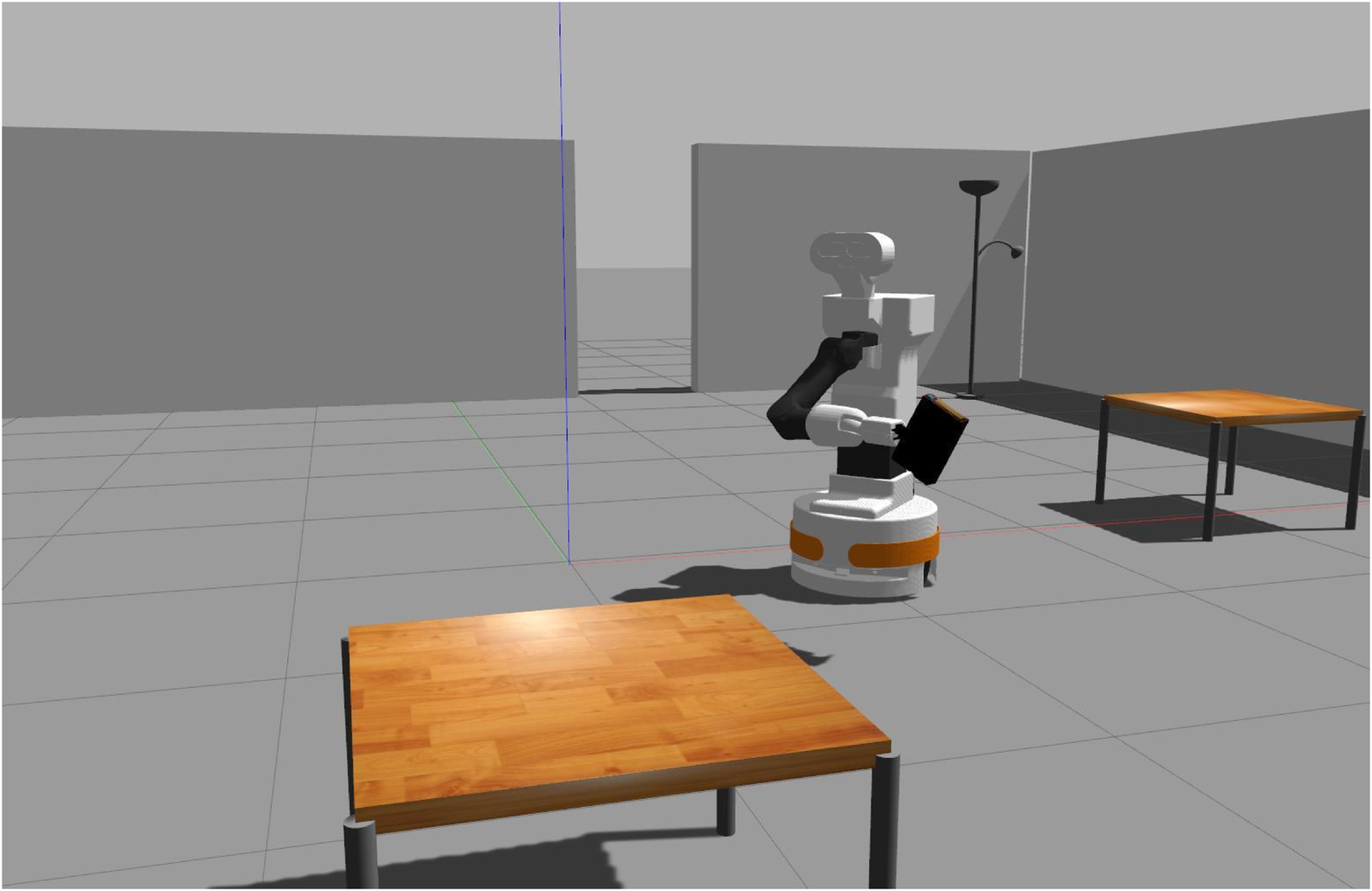}} &
 		\fbox{\includegraphics[width=0.32\columnwidth,trim={0.0cm 0.0cm 0.0cm 0.0cm},clip,angle =0]{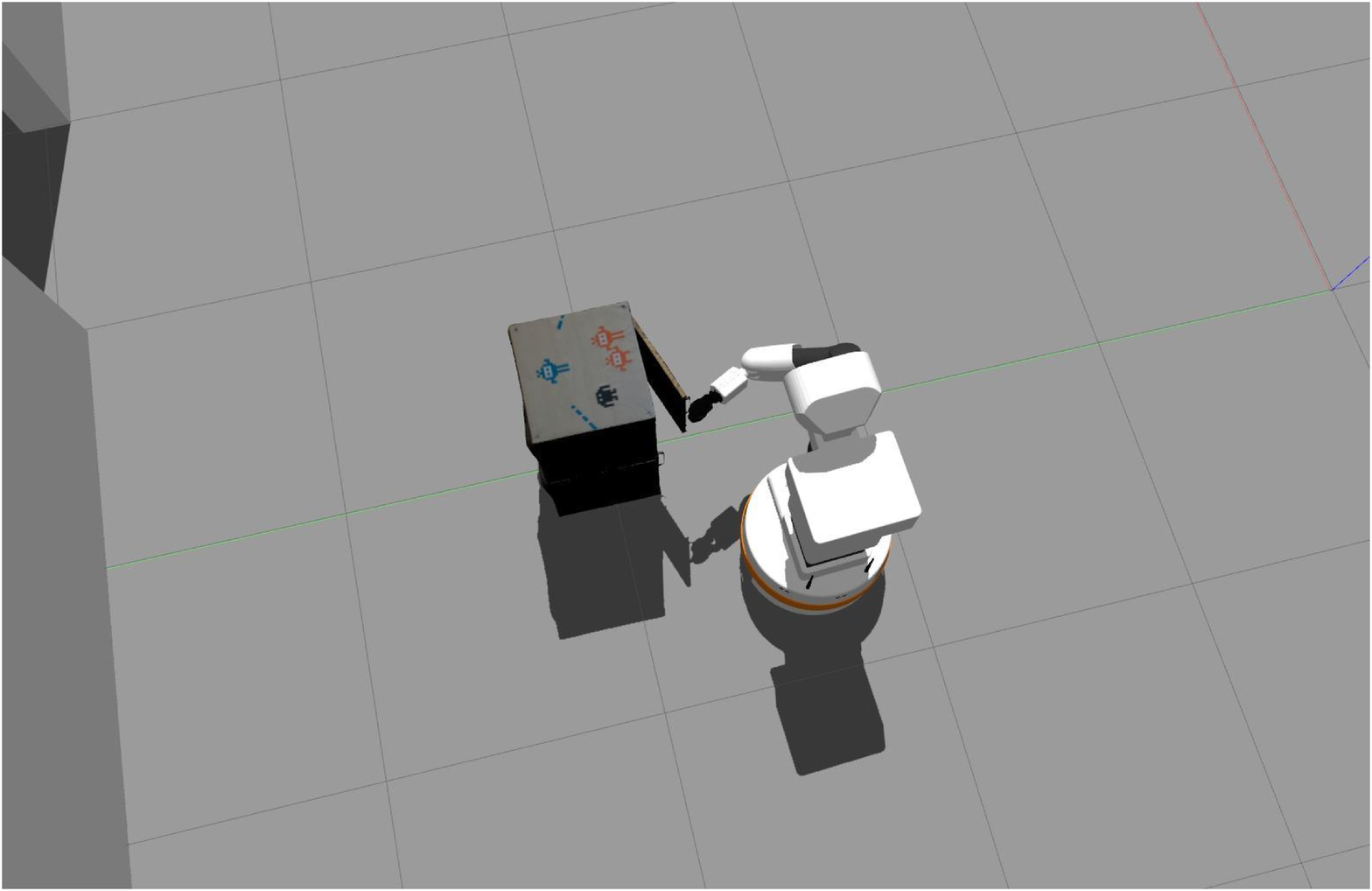}} &
 		\fbox{\includegraphics[width=0.32\columnwidth,trim={0.0cm 0.0cm 0.0cm 0.0cm},clip,angle =0]{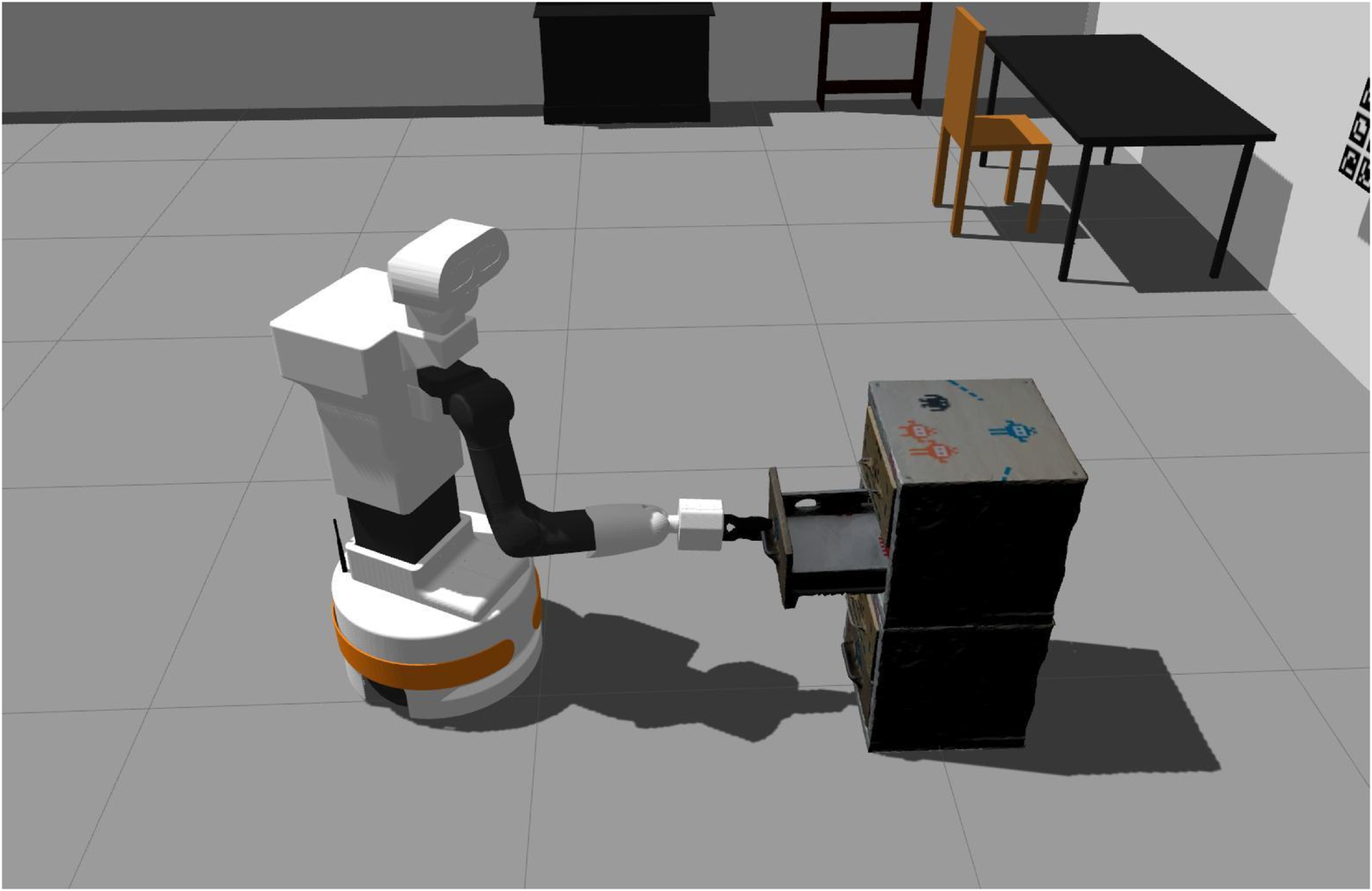}}\\
 		\fbox{\includegraphics[width=0.32\columnwidth,trim={0.0cm 0.0cm 0.0cm 0.0cm},clip,angle =0]{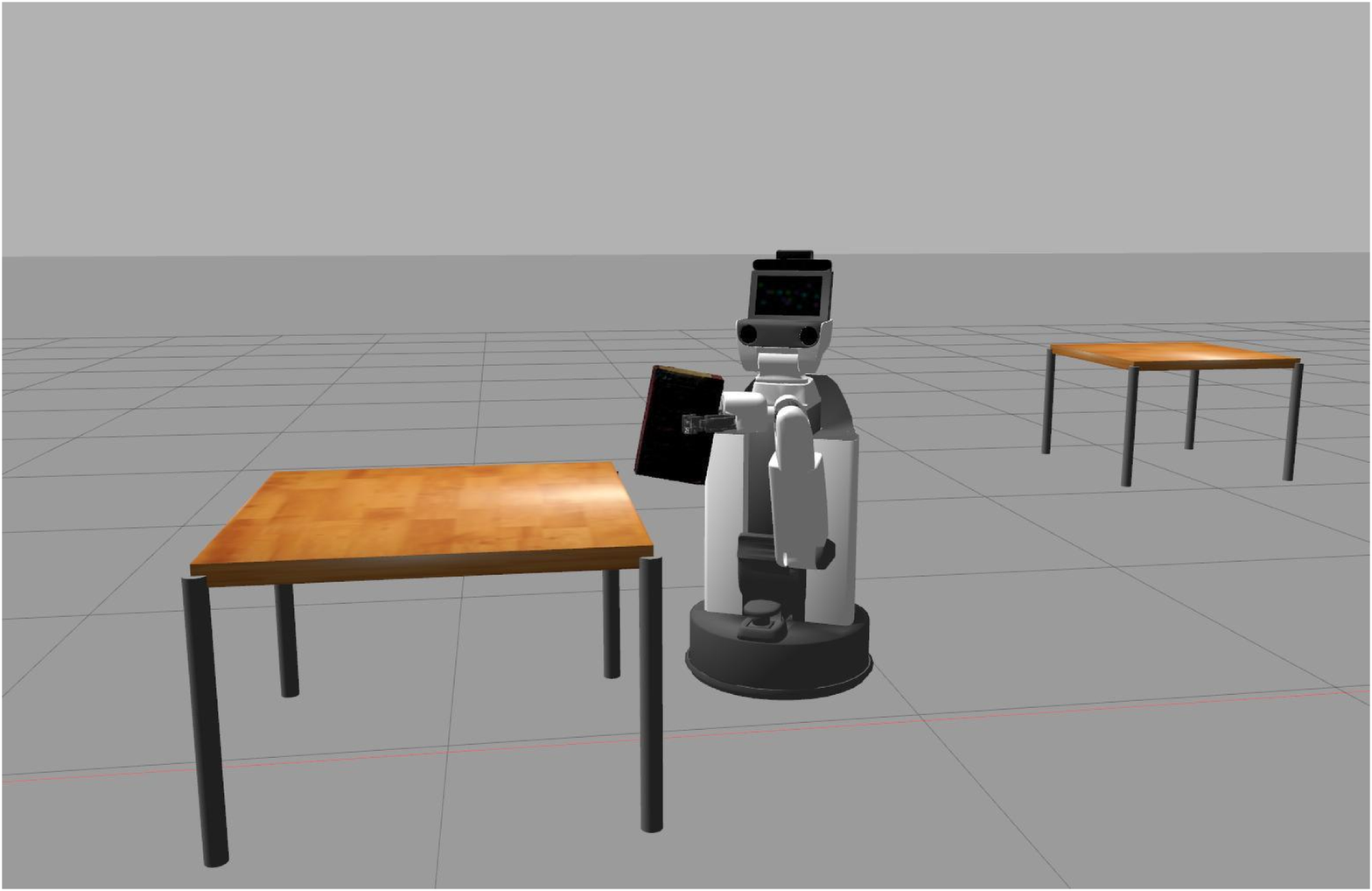}} &
 		\fbox{\includegraphics[width=0.32\columnwidth,trim={0.0cm 0.0cm 0.0cm 0.0cm},clip,angle =0]{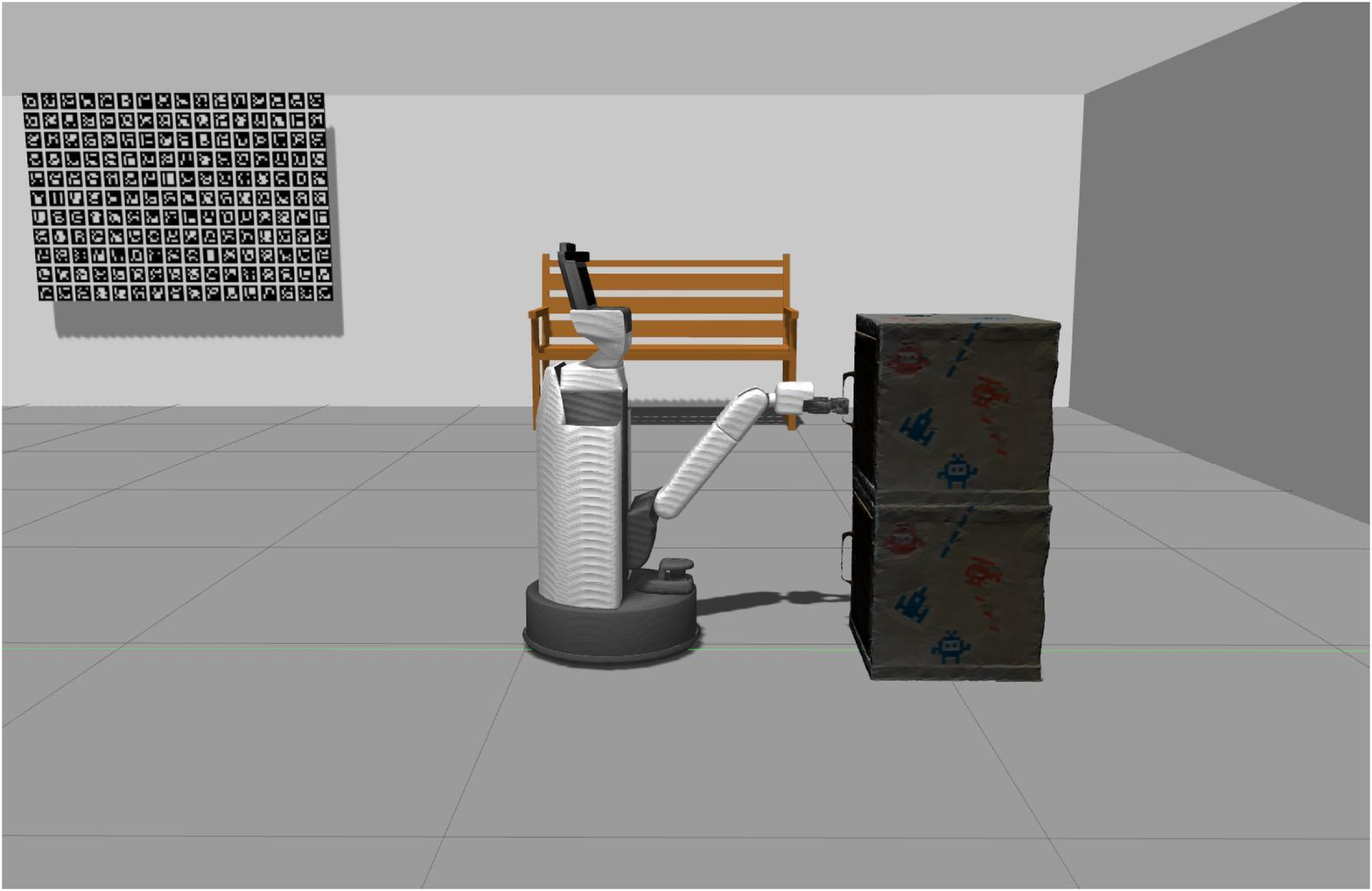}} &
 		\fbox{\includegraphics[width=0.32\columnwidth,trim={0.0cm 0.0cm 0.0cm 0.0cm},clip,angle =0]{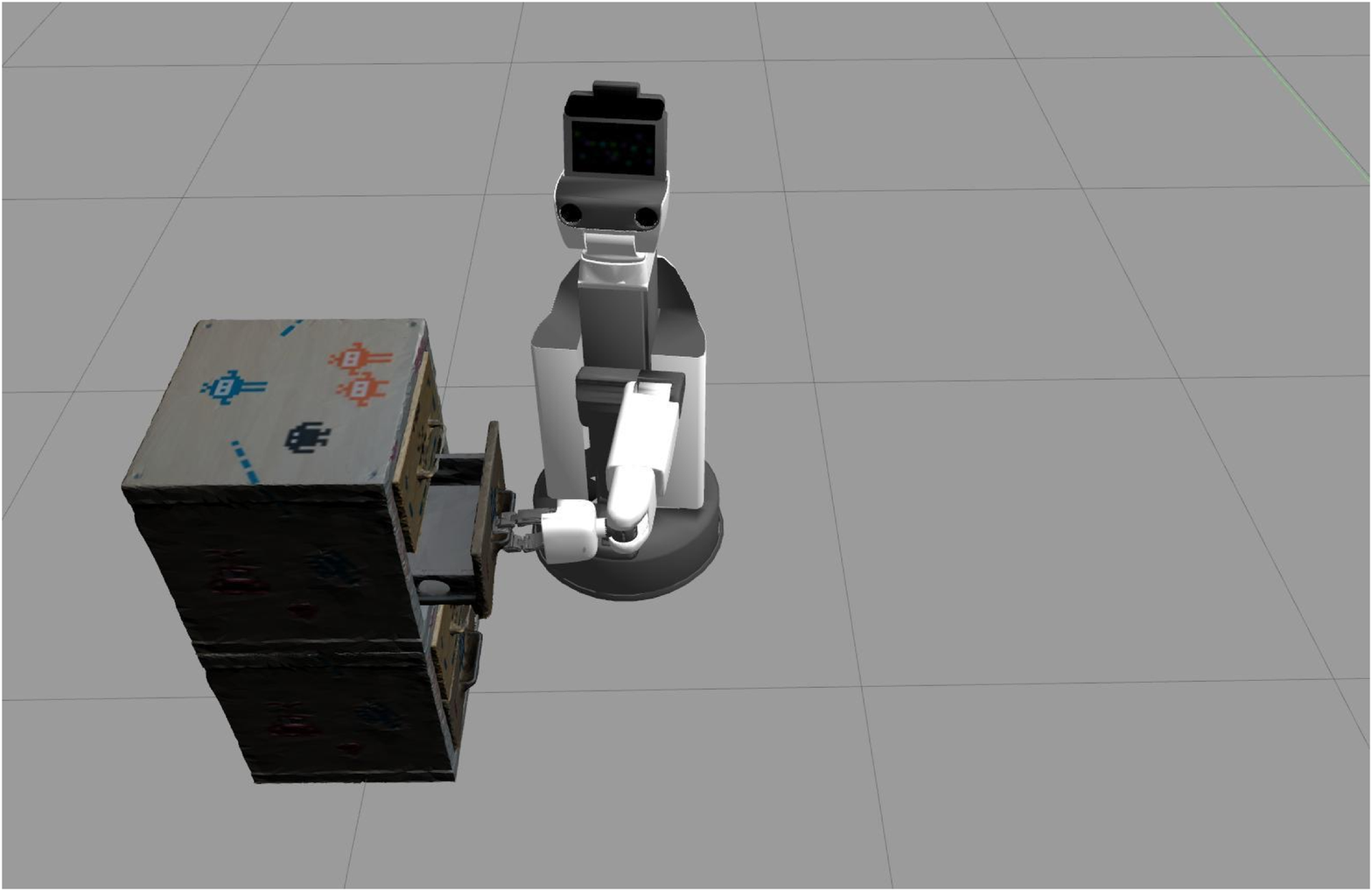}}
	\end{tabular}
	}
	\caption{Example trajectories in the Gazebo physics simulator on the pick\&place (left), door (mid) and drawer opening (right) tasks for the PR2 (top), TIAGo (mid) and the HSR (bottom) robots.}
  	\label{fig:gazebo}
\end{figure}

We further analyse how well the generated motions can be executed by running the system in the Gazebo simulator~\cite{Koenig-2004-394}. This includes both a simulation of the robot controllers as well as a physical simulation of the environment. We run all the agents at a frequency of $\SI{50}{\hertz}$. By still letting the agent observe the EE-velocities $\mathbf{v_{ee}}$ planned for a next timestep of $\SI{10}{\hertz}$, we can easily vary the frequency of the control loop without having to adapt the agent. The results across all approaches and tasks are summarized in \tabref{tab:gazebo} and visualized in \figref{fig:gazebo}. Differences to the analytical environment include previously potentially unmodelled accelerations, inertias and constraints, execution time of the arm movements as well as collisions with the physical objects. \blue{We do not tune any parameters of the low-level controllers to avoid to introduce any expert knowledge.}
We find particularly large inertias for the \blue{simulated} PR2 to cause the EE to outrun the base. To account for this, we slow down the EE-motions by a factor of two for all PR2 experiments in this section. Note that this was not needed in the real world experiments in \secref{sec:experiments_real}.

While all approaches experience a drop in performance compared to the analytical environment, we find this drop to be relatively small for the learned base motions, showing that they can be readily executed on all platforms. The average difference in performance across all tasks is 8.4\% for PR2, 8.0\%  for TIAGo and 3.0\%  for HSR. We identify two main causes for this:

\paragraph{Obstacles}
A main limitation of the current approach is that the base agent does not take the obstacles into account. 
To measure the impact of collisions with the physical objects, we execute the same motions with the objects removed from the scene, labelled \textit{nObj} in \tabref{tab:gazebo}. We find that this explains the largest part of the difference to the analytical environment for TIAGo, while the PR2 and HSR were not affected strongly by collisions. The gap to the analytical environment reduces to an average of 7.1\% for PR2, 2.6\% for TIAGo and 2.8\% for HSR. On average, the performance with physical objects is 2.3\% lower across all the tasks, further indicating that the learned base motions \sout{are economical and do not unnecessarily move or rotate around} \blue{avoid unnecessary movements or rotations}. Nonetheless, obstacle avoidance is an important piece to enable a wider variety of task setups. We aim to extend our formulation to incorporate this in future work.

\paragraph{Inertia}
For the PR2 we find that inertias in the base movement often cause failures in the beginning of the motion when the arm starts in an already stretched out position and then wants to move further away from the base. We extend our model to give the PR2 a "head start" of three seconds to position itself before we begin the EE-motion (\textit{PR2\_hs}). To do so, we use the same trained model, i.e. the agent did not see this head start during training. This simple adaptation closes the gap to the analytical environment to 2.6\%. Alternative approaches to account for large inertias would be to include acceleration limits in the linear motion system or to learn a scaling term to allow the agent to influence the EE-velocities.

The performance again further improves if we instead measure the share of episodes deviating less than $\SI{5}{\centi\meter}$ from the desired motions for both PR2 and TIAGo and the performance only slightly drops for the HSR. 

\setlength{\tabcolsep}{1pt}
\begin{table}
    \centering
    \begin{tabularx}{0.485\textwidth}{l|YYY|Y Z}
    \toprule
      Agent   & \multicolumn{3}{c|}{Linear Dynamic System} & \multicolumn{2}{c}{Imitation Learning}\\
      \cmidrule{2-6}
            & ggr & ggr restr & pick\&place & door & drawer \\
      \midrule
        PR2\_bl             & 48.2\hspace{1.0pt}(53.0) & 56.6\hspace{1.0pt}(62.2) & 35.4\hspace{1.0pt}(41.2) & \phantom{0}3.6 \phantom{0}(4.8) & \phantom{0}3.2\hspace{1.0pt}\phantom{0}(4.2)\\
        PR2\_gm             & 17.0\hspace{1.0pt}(22.2) & 19.4\hspace{1.0pt}(22.8)& 20.8\hspace{1.0pt}(24.6) & 21.2\hspace{1.0pt}(31.4) & \phantom{0}0.1\hspace{1.0pt}\phantom{0}(4.8)\\
        \textbf{PR2}        & \textbf{80.2\hspace{1.0pt}(83.6)} & \textbf{84.4\hspace{1.0pt}(88.8)} & \textbf{85.6\hspace{1.0pt}(88.8)} & \textbf{88.0\hspace{1.0pt}(92.8)} & \textbf{85.4\hspace{1.0pt}(91.6)}\\
        \textbf{PR2\_hs}    & \textbf{87.0\hspace{1.0pt}(88.0)} & \textbf{89.0\hspace{1.0pt}(90.8)} & \textbf{92.0\hspace{1.0pt}(93.6)} & \textbf{94.2\hspace{1.0pt}(95.0)} & \textbf{90.4\hspace{1.0pt}(90.6)}\\
        \textbf{PR2\_nObj}  & - \phantom{(xx.x)} & - \phantom{(xx.x)} & \textbf{87.6\hspace{1.0pt}(92.4)} & \textbf{85.6\hspace{1.0pt}(90.0)} & \textbf{85.4\hspace{1.0pt}(90.8)}\\
        \cmidrule{1-6}
        Tiago\_bl           & \phantom{0}7.0\hspace{1.0pt}\phantom{0}(7.8) & \phantom{0}9.0\hspace{1.0pt}(11.0) & \phantom{0}2.2\hspace{1.0pt}\phantom{0}(2.8) & \phantom{0}1.6\hspace{1.0pt}\phantom{0}(3.0) & \phantom{0}6.8\hspace{1.0pt}\phantom{0}(7.8)\\
        \textbf{Tiago}      & \textbf{65.4\hspace{1.0pt}(70.4)} & \textbf{74.6\hspace{1.0pt}(81.2)} & \textbf{88.2\hspace{1.0pt}(91.2)} & \textbf{81.6\hspace{1.0pt}(86.8)} & \textbf{83.8\hspace{1.0pt}(88.0)}\\
        \textbf{Tiago\_nObj}         & - \phantom{(xx.x)} & - \phantom{(xx.x)} & \textbf{87.4\hspace{1.0pt}(91.0)} & \textbf{93.4\hspace{1.0pt}(95.4)} & \textbf{92.8\hspace{1.0pt}(94.2)}\\
        \cmidrule{1-6}
        HSR\_bl             & \phantom{0}2.6\hspace{1.0pt}\phantom{0}(0.0) & \phantom{0}2.6\hspace{1.0pt}\phantom{0}(0.1) & \phantom{0}0.0\hspace{1.0pt}\phantom{0}(0.0) & \phantom{0}0.0\hspace{1.0pt}\phantom{0}(0.0) & \phantom{0}0.0\hspace{1.0pt}\phantom{0}(0.0)\\
        \textbf{HSR}        & \textbf{64.4\hspace{1.0pt}(54.0)} & \textbf{70.3\hspace{1.0pt}(59.4)} & \textbf{85.6\hspace{1.0pt}(69.6)} & \textbf{87.2\hspace{1.0pt}(78.0)} & \textbf{90.2\hspace{1.0pt}(83.8)}\\
        \textbf{HSR\_nObj}           & - \phantom{(xx.x)} & - \phantom{(xx.x)} & \textbf{92.2\hspace{1.0pt}(78.2)} & \textbf{87.2\hspace{1.0pt}(76.6)} & \textbf{87.4\hspace{1.0pt}(82.4)}\\
        \bottomrule
    \end{tabularx}
    \caption{Performance in the Gazebo physics simulator as share of successfully executed episodes with zero kinematically infeasible EE-poses and share of episodes that never deviate more than $\SI{5}{\centi\meter}$ from the EE-motion (in brackets). The bold rows represent our proposed approach \blue{and two modifications thereof for identification of error sources}.}
    \label{tab:gazebo}
\end{table}
\setlength{\tabcolsep}{6pt}

\subsection{Motion Execution in Real World Setting }
\label{sec:experiments_real}

\begin{figure}
	\centering
	\resizebox{\columnwidth}{!}{%
  	\begin{tabular}{ccc}
 		\fbox{\includegraphics[width=0.32\columnwidth,trim={1.25cm 0.625cm 4.75cm 1.8625cm},clip,angle =0]{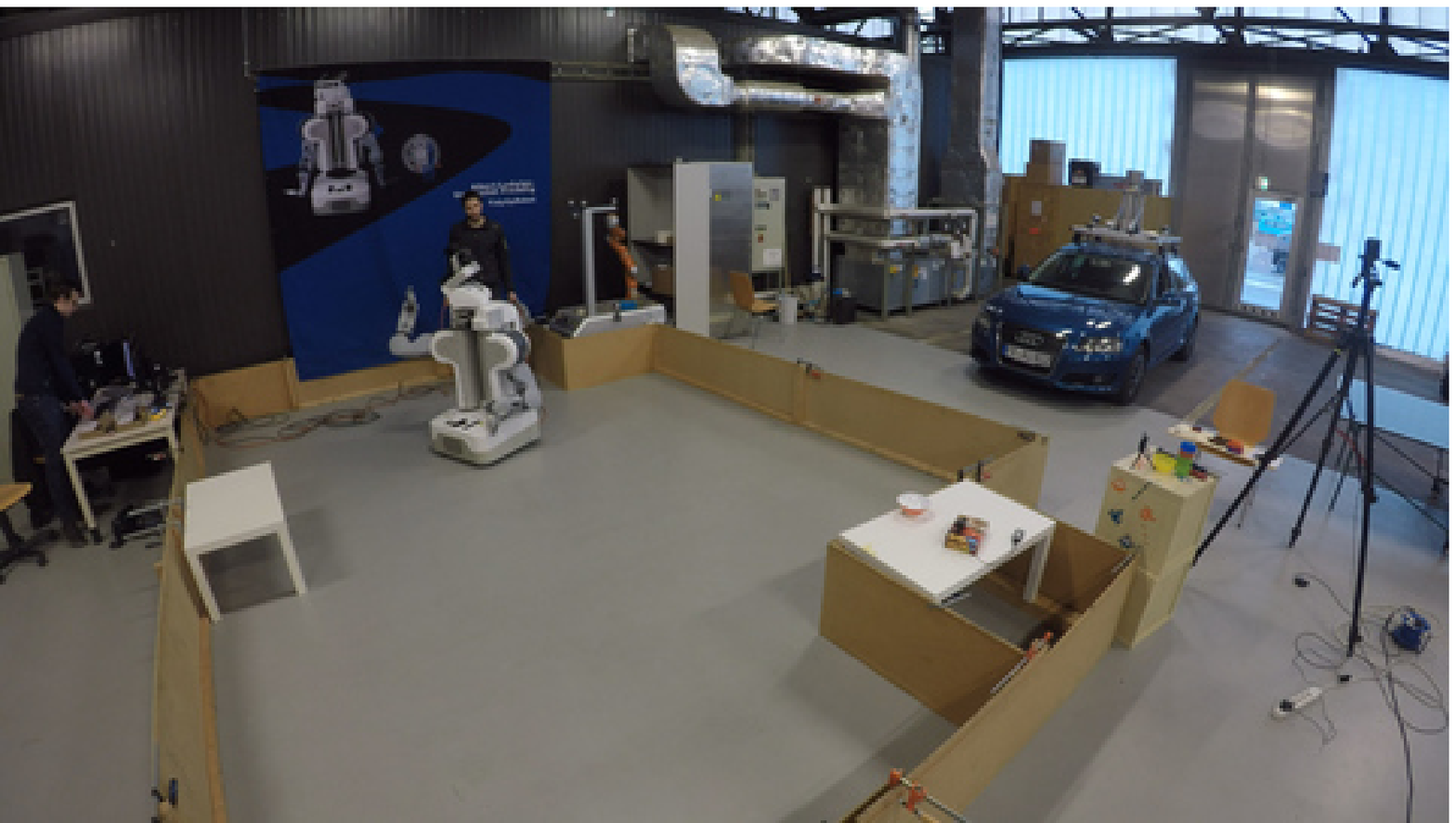}} &
 		\fbox{\includegraphics[width=0.32\columnwidth,trim={1.25cm 0.625cm 4.75cm 1.8625cm},clip,angle =0]{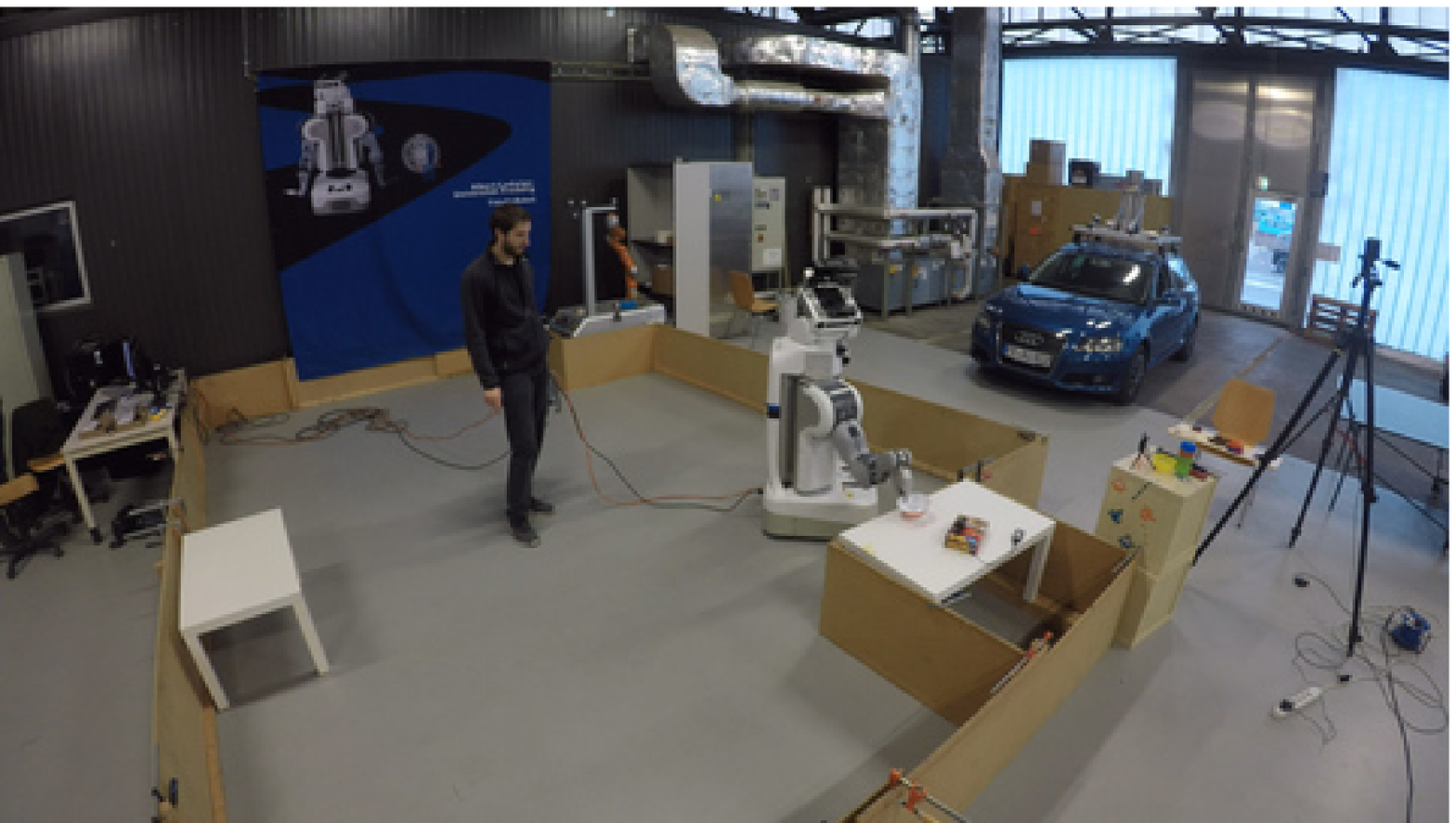}} &
 		\fbox{\includegraphics[width=0.32\columnwidth,trim={1.25cm 0.625cm 4.75cm 1.8625cm},clip,angle =0]{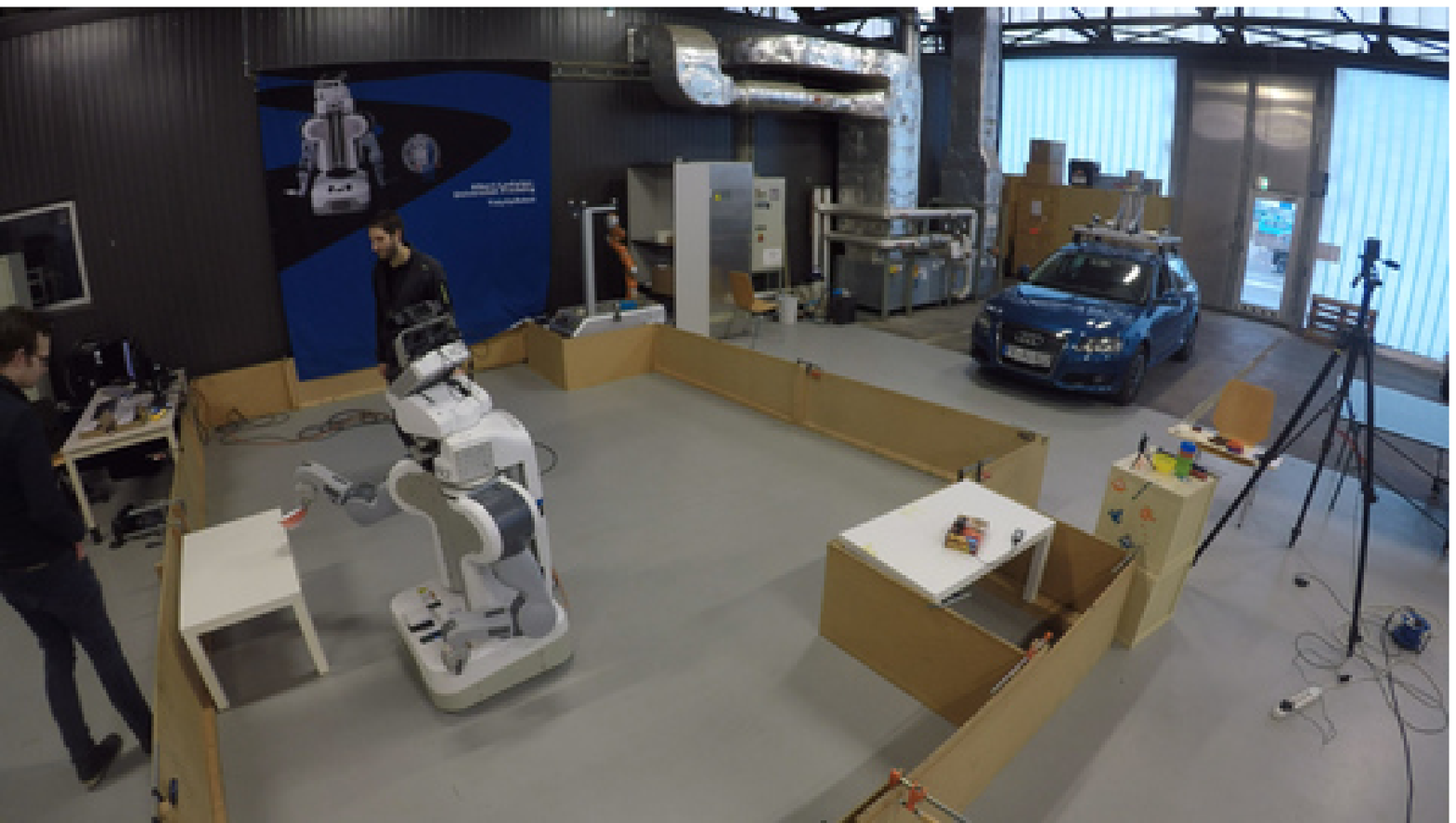}}\\
  		\fbox{\includegraphics[width=0.32\columnwidth,trim={0.0cm 1.25cm 3.5cm 0.0cm},clip,angle =0]{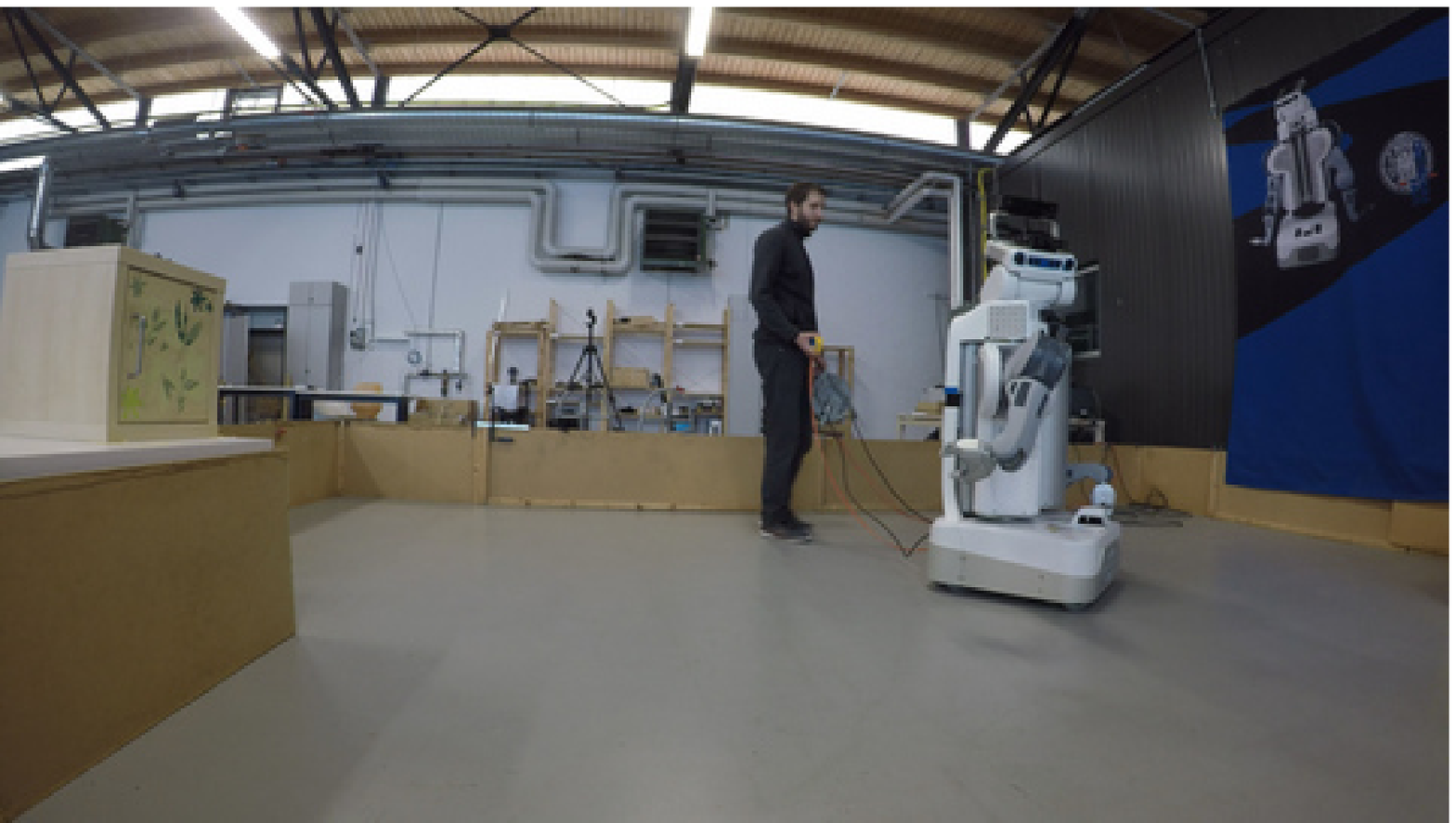}} &
 		\fbox{\includegraphics[width=0.32\columnwidth,trim={0.0cm 1.25cm 3.5cm 0.0cm},clip,angle =0]{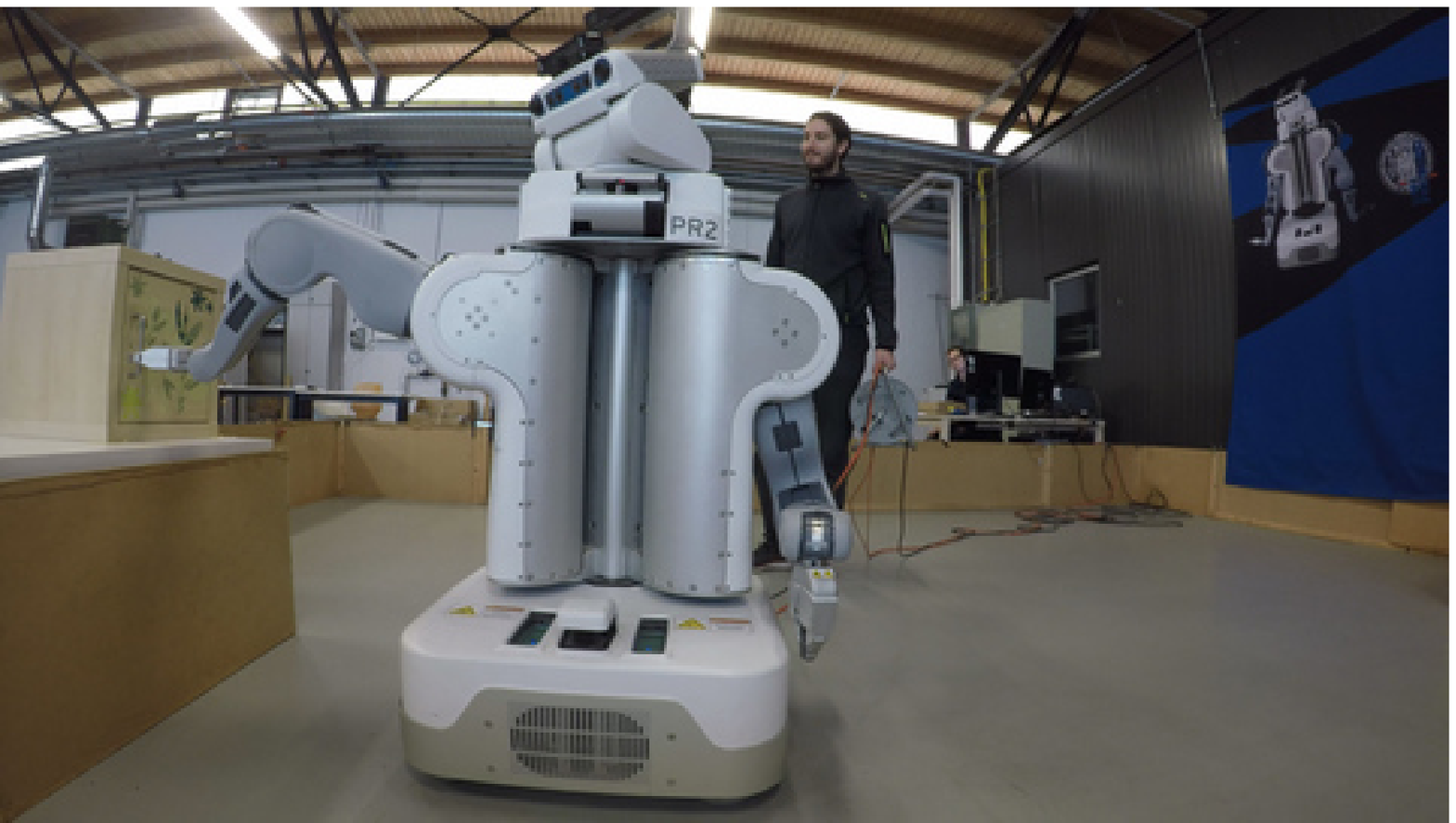}} &
 		\fbox{\includegraphics[width=0.32\columnwidth,trim={0.0cm 1.25cm 3.5cm 0.0cm},clip,angle =0]{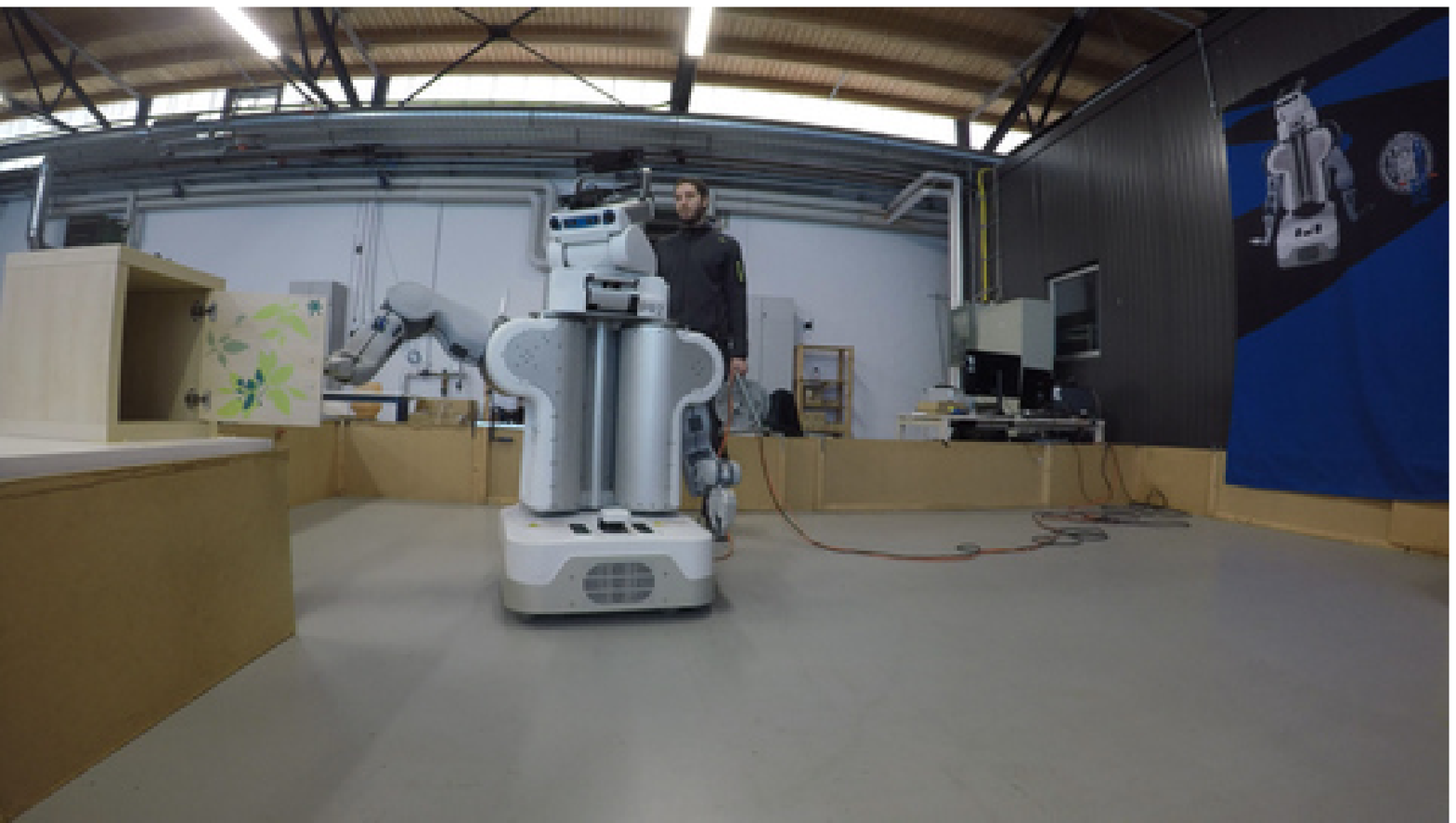}}\\ 
 	    \fbox{\includegraphics[width=0.32\columnwidth,trim={0.0cm 0.5cm 2.0cm 0.0cm},clip,angle =0]{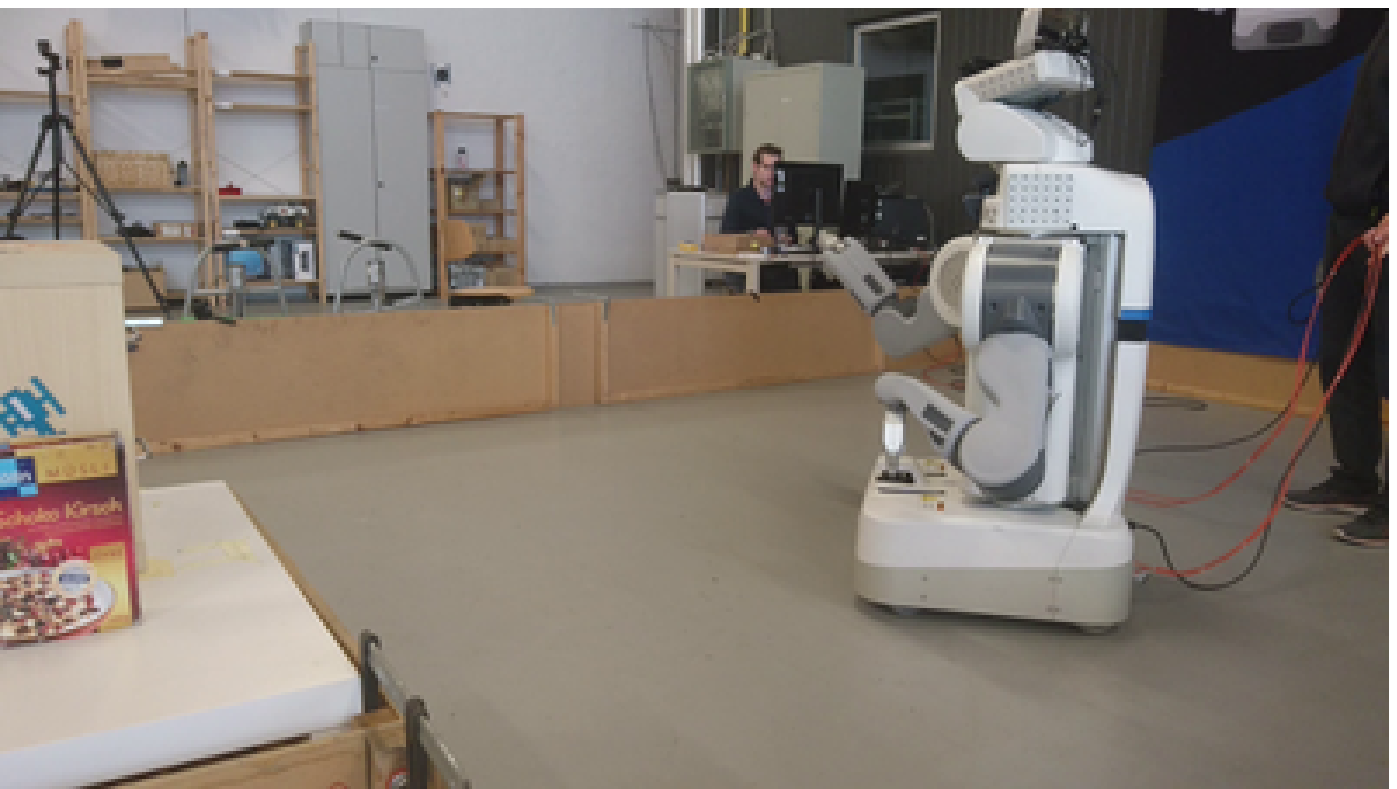}} &
  		\fbox{\includegraphics[width=0.32\columnwidth,trim={0.0cm 0.5cm 2.0cm 0.0cm},clip,angle =0]{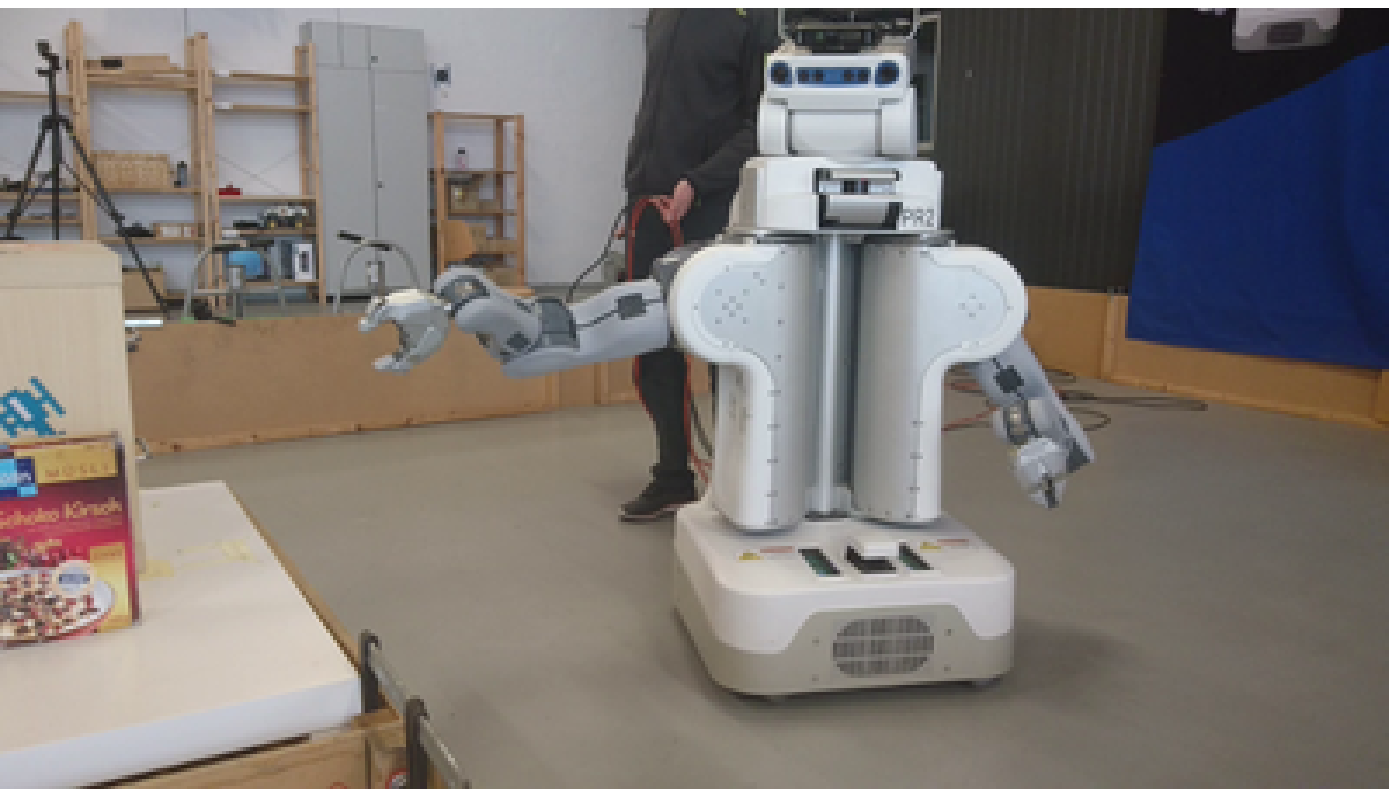}} &
  		\fbox{\includegraphics[width=0.32\columnwidth,trim={0.0cm 0.5cm 2.0cm 0.0cm},clip,angle =0]{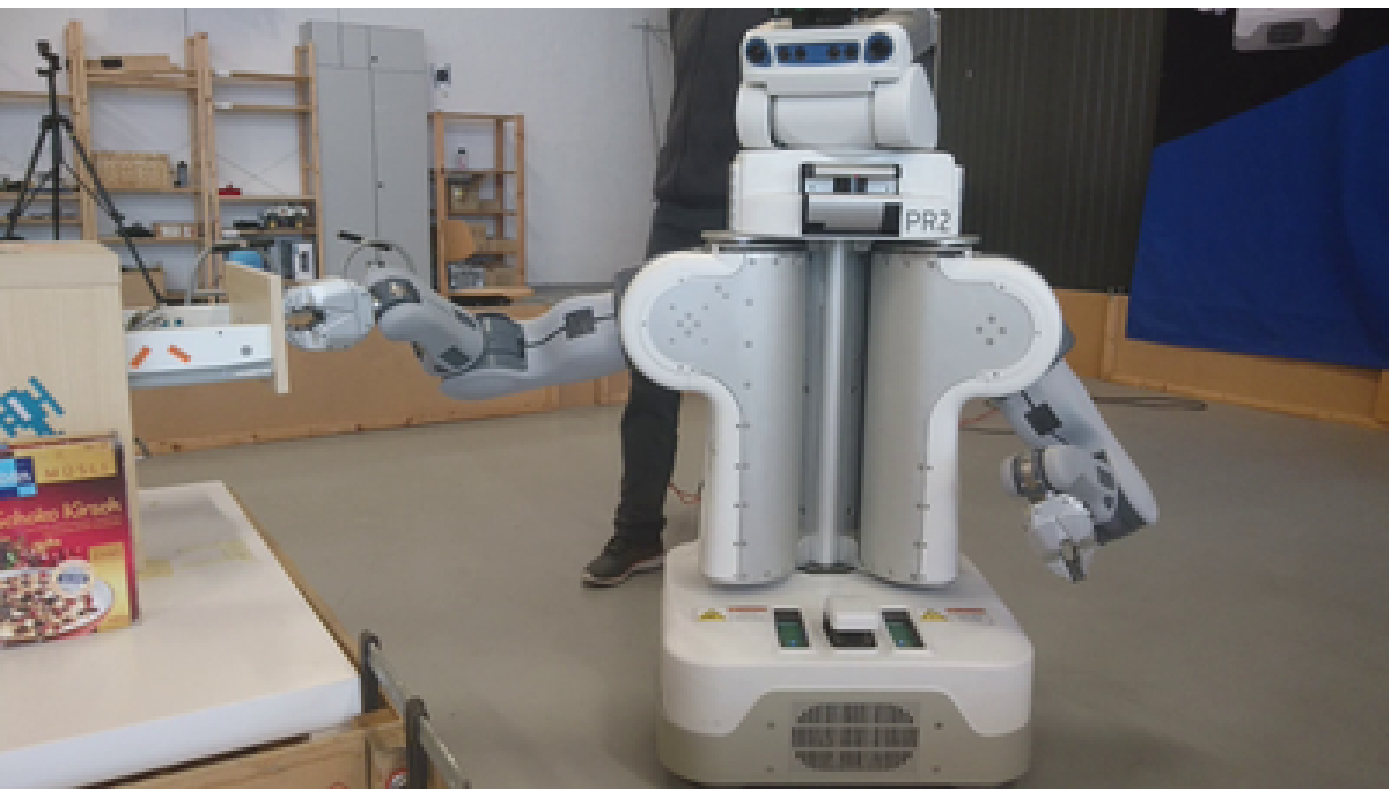}}
	\end{tabular}
	}
	\caption{Snapshots of action execution on the PR2 robot. Given the position of the manipulated objects the robot performs the \textit{pick\&place}, \textit{grasp and open a cabinet door}  and \textit{grasp and open a drawer} tasks (top to bottom).}
  	\label{fig:PR2real}
\end{figure}

To further demonstrate the applicability of our approach in real world settings, we evaluate all the described tasks on a PR2. We construct a small environment of roughly $\SI{4.5}{\meter}$ $\times$ $\SI{6}{\meter}$ which is shown in \figref{fig:PR2real}. \blue{The PR2 uses its base scanner and a standard implementation of Adaptive Monte Carlo Localization (AMCL) to localize itself in a static map. The poses of the target objects in map frame are provided manually.} To account for space constraints, we make the following adaptations:
(i) we start each episode of the \textit{general goal reaching} task from the last achieved base pose and only sample a new arm joint configuration; (ii) we reject random goals that lie outside the designated map minus a safety distance from the edges, resulting in a maximum distance of five meters between consecutive goals; (iii) in \textit{pick\&place}, we randomise the starting pose and drop off location but leave the pickup location fixed.

As in the analytical and simulated experiments, we only count executions without a single failure as kinematic success. This may include episodes where the grasp of the manipulated object fails (this does not lead to interruption).
Additionally, we report episodes as task success if the manipulation is executed as desired. This may include episodes with up to $99$ kinematic failures at which point we interrupt task execution. This corresponds to roughly two seconds of unsuitable configurations ($50$Hz control). The results are shown in \tabref{tab:real} and exemplary snapshots from the task execution are shown in \figref{fig:PR2real}. We achieve good overall results with an average of $87.5\%$ kinematic success in the task execution. \blue{All episodes were completed without a collision or human intervention such as emergency stops. This aligns with the results from Gazebo, further indicating that the agent learns to avoid unnecessarily extensive base movements.} Nonetheless we identified several sources of error for further improvement:
\paragraph{Controller Issues}
Throughout all experiments we control the motion of robot arm and base independently. In phases of fast base rotation the arm controller sometimes lags behind in compensating the rotary motion of the base. This can lead to grasp failures or cause kinematic failures in subsequent time steps. \blue{However, this is a result of the independence of the low level base and arm controllers and not caused by our approach. A combined controller for robot arm and base could be integrated without requiring any adaptations.}
\paragraph{Unusual Starting Configurations}
Most kinematic failures occur at beginning of trajectories and are caused by difficult starting configurations and the restrictive \sout{end-effector planner}\blue{EE-motion generator}. There are very few kinematic failures during manipulation itself, i.e. when the arm is in a typical working configuration.
\paragraph{Configuration Jumps} Occasionally the solutions found by the IK-solver jump to very different arm joint configurations from one step to the next. These can lead to grasp failure and kinematic problems as it causes the gripper motion to deviate from the desired trajectory as the configuration shift is not instantaneous as in analytical training.

\setlength{\tabcolsep}{1pt}
\renewcommand{\arraystretch}{1}
\setlength{\tabcolsep}{6pt}
\renewcommand{\arraystretch}{1}
 
Overall the policy shows to be robust to real-world noise and inertia as well as control reaction times. Demonstrating further that the base policy learns to seek robust behaviours and positioning and successfully transfers to real-world settings in a zero-shot manner.

\setlength{\tabcolsep}{4pt}
\begin{table}
    \centering
    \begin{tabularx}{0.485\textwidth}{l|YY|Y Y}
    \toprule
      PR2   & \multicolumn{2}{c|}{Linear Dynamic System} & \multicolumn{2}{c}{Imitation Learning}\\
      \cmidrule{2-5}
      & ggr & pick\&place & door & drawer \\
      \midrule
        Kinematic success         & 46 & 45 & 41 & 43\\
        Task success              & -  & 48 & 43 & 40\\
        Total episodes       & 50 & 50 & 50 & 50\\
      \bottomrule
    \end{tabularx}
    \caption{Performance on the real world PR2 robot as number of episodes. Kinematic success refers to episodes with zero kinematic failures, task success to the completion of the task objective (e.g. door opened).}
    \label{tab:real}
\end{table}
\setlength{\tabcolsep}{6pt}

\setlength{\tabcolsep}{1pt}
\setlength{\tabcolsep}{6pt}

\section{Conclusions}

In this paper, we presented an approach to generate suitable robot base motions for mobile platforms given an end-effector motion generated by an arbitrary system. We formulated the problem as a reinforcement learning setting in which the robot configuration, the end-effector velocity and goal serve as observations and the robot base velocities are the corresponding actions. The environment reward is derived from the kinematic feasibility of the resulting robot base and end-effector poses. Leveraging state-of-the-art RL methods we achieve high success rates across different robot platforms for both seen and unseen end-effector motions. This demonstrates the potential of this approach to enable the application of any system that generates task specific end-effector motions across a diverse set of mobile manipulation platforms.

While we achieve very good results in terms of kinematic feasibility during trajectory generation, the approach so far does not consider collisions with the environment. In future work, we plan to incorporate obstacles and object detection into the training to enable the agent to also avoid collision while moving the base. Furthermore, we observed occasional undesired configuration jumps in the arms with high degrees of freedom which we aim to avoid by incentivising smooth joint movements through the reward function.
In order to improve the performance in real world deployment, we will investigate the use of a shared controller for base and arm that will allow a faster and smoother compensation of the arm towards motion of the base. In addition to yielding a higher accuracy, this will also allow for faster execution.



\footnotesize
\bibliographystyle{IEEEtran}
\bibliography{icra21}


\end{document}